\documentclass[11pt]{article}
\usepackage{amssymb}
\usepackage[final]{acl}
\usepackage{amsmath}
\usepackage{times}
\usepackage{latexsym}
\usepackage{longtable}
\usepackage{csquotes}
\usepackage{booktabs}
\usepackage{algorithm}
\usepackage{algpseudocode}
\usepackage{placeins}
\usepackage{enumitem}
\usepackage{tikz}
\usetikzlibrary{shapes.geometric, arrows.meta, positioning, fit, calc, shadows.blur, patterns, decorations.pathmorphing}
\usepackage{fontawesome5} % 需要加载这个包来显示放大镜、齿轮等小图标
\usepackage[T1]{fontenc}
\usepackage[utf8]{inputenc}

\usepackage{microtype}

\usepackage{inconsolata}
\usepackage{multirow}
\usepackage{graphicx}

\title{Synthetic or Authentic? Building Mental Patient Simulators from Longitudinal Evidence}

\author{
Baihan Li$^{1,2}$,
Bingrui Jin$^{1,2}$,
Kunyao Lan$^{2}$,
Ming Wang$^{3,4}$,
Mengyue Wu$^{2,*}$ \\
$^{1}$SJTU Paris Elite Institute of Technology, Shanghai Jiao Tong University, China \\
$^{2}$X-LANCE Lab, School of Computer Science, Shanghai Jiao Tong University, Shanghai, China \\
$^{3}$School of Computer Science and Engineering, Northeastern University, China \\
$^{4}$School of Computing and Information Systems, Singapore Management University, Singapore \\
\texttt{lbh0612@sjtu.edu.cn, mengyuewu@sjtu.edu.cn} \\
\small{$^{*}$Corresponding author}
}

\begin{document}
\maketitle

\begin{abstract}
Patient simulation is essential for developing and evaluating mental health dialogue systems. As most existing approaches rely on snapshot-style prompts with limited profile information, homogeneous behaviors and incoherent disease progression in multi-turn interactions have become key challenges.
In this work, we propose \textsc{Deprofile}, a data-grounded patient simulation framework that constructs unified, multi-source patient profiles by integrating demographic attributes, standardized clinical symptoms, counseling dialogues, and longitudinal life-event histories from real-world data. We further introduce a Chain-of-Change agent to transform noisy longitudinal records into structured, temporally grounded memory representations for simulation. Experiments across multiple large language model (LLM) backbones show that with more comprehensive profile constructed by \textsc{Deprofile}, the dialogue realism, behavioral diversity, and event richness have consistently improved and exceed state-of-the-art baselines, highlighting the importance of grounding patient simulation in verifiable longitudinal evidence\footnote{Repository with code, data structure descriptions, and evaluation
% scripts is available at \url{https://anonymous.4open.science/r/Deprofile}.
scripts is available at \url{https://github.com/Baihan-12/Deprofile}
}
\end{abstract}

% =========================
\section{Introduction}

% 【TODO】
% 1. 抑郁症患者模拟的研究背景与应用场景
% 2. 现有 patient agent 的主要局限（静态 persona / prompt-level）
% 3. 强调“时间演化 + 生活事件 + 咨询上下文”的现实重要性
% 4. 引出本文的 unified patient profile
% 5. 列出 contribution（3–4 点）
% \paragraph{}
The simulation of patients with mental disorders holds significant potential for advancing psychiatric training, therapy development, and the automated evaluation of diagnostic and therapeutic models. Accurate patient simulators can serve as safe, scalable environments for clinicians to practice interventions and for researchers to test computational models of care. However, this task presents a profound challenge.
While Large Language Models (LLMs) have demonstrated remarkable potential in simulating human cognition and behavior, existing approaches are predominantly ``snapshot-based''~\citep{liu2025eeyore,louie2024roleplay,schick2024talkdep,li2025patientsim}. They generate responses from static biographical tags or short conversational contexts, creating fluent but ultimately superficial interactions. This paradigm overlooks the essential clinical reality that mental illness is not a static condition but a \textbf{dynamic, longitudinal process}, a lived experience shaped by a temporal stream of life events, fluctuating symptoms, and evolving internal states~\citep{Yarrington2023LifeEvents,Maike2022StressfulLifeEvents}. Capturing this temporal depth is crucial for simulations that are not just linguistically plausible
but clinically meaningful and practically useful for downstream applications in training and automated evaluation.

\begin{figure*}[t] % 加星号表示跨双栏，[t] 表示固定在页面顶部
    \centering
    % 使用 \textwidth 确保图片铺满左右两栏的宽度
    \includegraphics[width=0.9\textwidth]{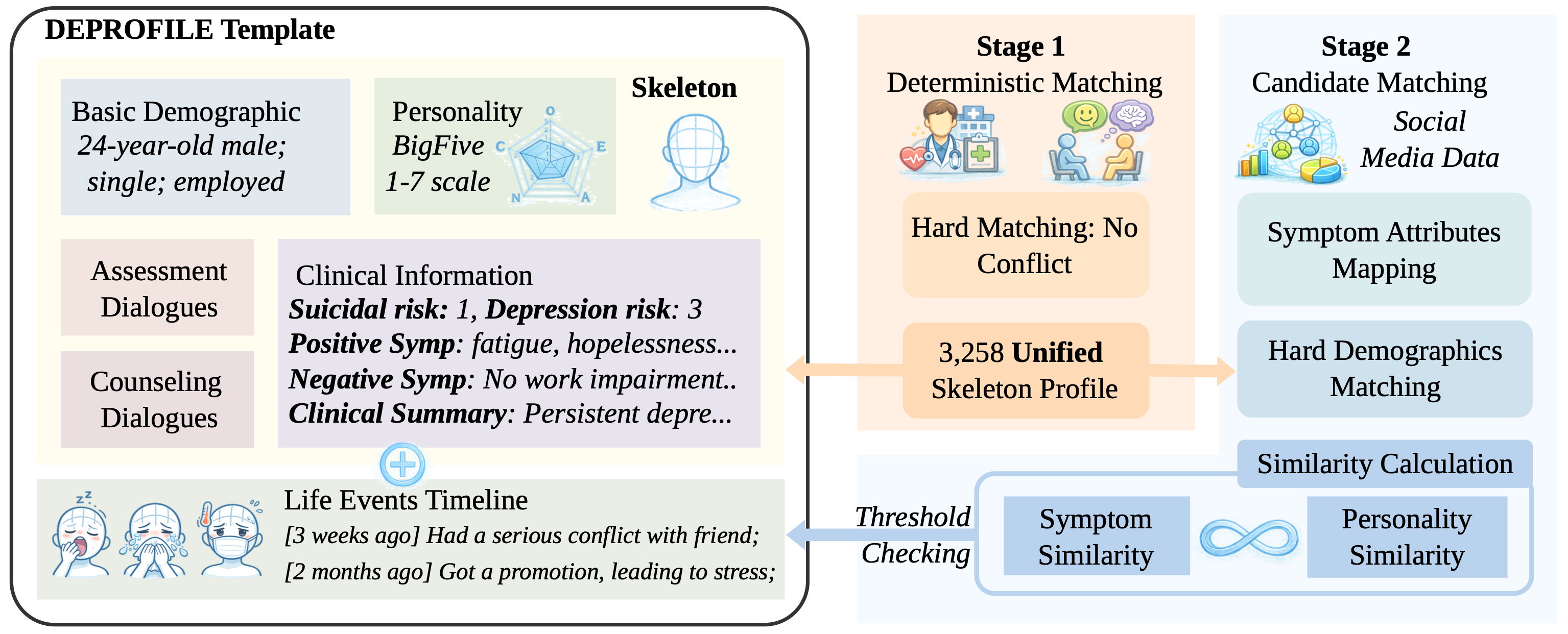} 
    
    % 这里是为你量身定制的高级标题
    \caption{The \textsc{Deprofile} Framework Overview}
    
    \label{fig:deprofile_full} % 换个更有辨识度的标签名
\end{figure*}

% \begin{figure}[htbp]
%     \centering
%     % 调整宽度：0.9\textwidth 表示占页面宽度的 90%
%     \includegraphics[width=0.7\columnwidth]{latex/figures/picture1.png}
    
%     % 图注：ACL 风格通常要求首句概括全图，随后解释细节
%     \caption{ Overview of the \textsc{Deprofile} framework and its data composition.
% We construct a multifaceted depression patient simulator using real-world datasets,
% including assessment dialogues, counseling interactions, and longitudinal records of
% symptoms and life events.
%  }
    
%     % 引用标签
%     \label{fig:framework}
% \end{figure}

% Current models \MY{usually use xxxxx, you should have a shorterned review here, saying previous methods have done xx xx, yet lack. you need to highlight current models lack diversity since this is your core elevation} lack explicit modeling of this life event history, resulting in simulated patients behaving like actors without a past who fail to demonstrate logical disease progression during multi-turn interviews; \MY{here you can say most models rely heavily on LLM's innate knowledge by providing profile in concise prompts...then stress that it's data-scarce.}models often rely on their internal parametric knowledge to ``confabulate'' patient experiences. This leads not only to inconsistencies but also to the generation of stereotypical, non-clinical narratives, which severely undermines the seriousness of medical simulation.

Current patient simulation models typically construct personas using concise, snapshot-style prompts that specify a limited set of symptoms, stressors, or personality tags, and rely on the LLM to extrapolate a complete patient narrative from its internal knowledge.
While effective for generating fluent responses, this paradigm lacks explicit modeling of longitudinal life-event history and tends to produce homogeneous behaviors across profiles, resulting in simulated patients who behave like actors without a past and fail to exhibit coherent disease progression in multi-turn interviews.
Because these approaches are largely data-scarce and inference-driven, models often confabulate patient experiences based on parametric priors rather than grounded evidence, leading to internal inconsistencies and  stereotypical narratives that undermine the clinical realism.

To address these issues, we propose \textsc{Deprofile}, the first patient simulation framework driven by multi-source real-world data with longitudinal temporal depth.
Unlike previous methods relying solely on isolated dialogue data, we innovate a \textbf{two-stage matching mechanism} to align clinical \textbf{Assessment} interviews, \textbf{Counseling} dialogues, and users' longitudinal \textbf{Social Media} histories. 
Through this approach, we construct a multidimensional patient profile that contains not only precise descriptions of clinical symptoms but also life event trajectories spanning months or even years.

To tackle the noise and unstructured nature of social media data, we further design the \textbf{Chain-of-Change (CoC) Agent}. 
This agent extracts key ``life events'' and ``symptom fluctuations'' from chaotic social streams and transforms them into structured, retrievable \textbf{Memory Cards}. 
During generation, we implement a strict \textbf{dual-channel memory mechanism}: the model is constrained to answer questions about past experiences based on retrieved ``Memory Cards,'' thereby fundamentally suppressing the generation of hallucinations.

The main contributions of this paper are summarized as follows:
% \begin{itemize}[leftmargin=*,noitemsep,topsep=0pt]
%     \item We propose the \textsc{Deprofile}  framework, which bridges the data silos of diagnosis, counseling, and social media for the first time. We construct a unified patient profile library containing 3,258 high-quality samples, enabling dynamic simulation with real-world temporal awareness.
%     \item We design the CoC Agent and a structured memory extraction pipeline. This framework effectively converts unstructured longitudinal behavioral data into verifiable clinical evidence while suppressing hallucinations through a novel citation generation protocol.
%     \item Our experiments reveal a \textbf{``Dual Regulatory Mechanism''} of timeline information: it acts as a ``generative seed'' to enrich details in sparse contexts, while serving as a ``constraint anchor'' to ensure factual consistency in rich contexts. Furthermore, we demonstrate that under this framework, smaller open-source models can achieve clinical fidelity comparable to closed-source large models.

% \end{itemize}

\begin{itemize}[leftmargin=*,leftmargin=*,noitemsep,topsep=0pt]
    \item We propose \textsc{Deprofile}, the first framework bridging diagnostic, counseling, and social media silos. We construct a unified library of \textbf{3,258} high-quality profiles, enabling dynamic simulations with real-world temporal awareness and long-term behavioral consistency.
    \item We design the \textbf{CoC Agent} and a structured memory pipeline to convert unstructured longitudinal data into verifiable clinical evidence, effectively suppressing hallucinations through a novel citation generation protocol.
    \item We identify a \textbf{``Dual Regulatory Mechanism''}: timeline information acts as a ``generative seed'' to enrich sparse contexts and a ``constraint anchor'' to ensure factual consistency. This mechanism enables smaller open-source models to match proprietary LLMs in clinical fidelity.
\end{itemize}

\section{\textsc{Deprofile} : Unified Patient Profile}
%\paragraph{}
% This unified patient profile structure features 7 attributes by aligning three types of patient-related data: assessment interviews, counseling dialogues, and social media posts from mental-disordered users.
% , thereby representing depression patients across multiple perspectives and temporal scales.
%In this section, we first introduce the datasets used in our study, then describe how additional annotations and information extraction are performed for each data source, and finally explain how we construct unified patient-centric profiles with multiple candidate life event histories.

We construct unified profiles by aligning clinical interviews, counseling dialogues, and longitudinal social media records. In this section, we will: (1) introduce the three data sources used in our study; (2) explain the information extraction protocols and annotation strategies; and (3) describe the consistency-driven alignment that implements a two-stage matching mechanism to ensure a high-fidelity, unified profile library.

\subsection{Data Sources}

% \paragraph{}
% In this section, we describe the datasets used in our study.

\paragraph{Assessment Dialogue Data}
provides primary symptom information and chief complaints grounded in standardized clinical interviews.
We use the D4 dataset~\citep{yao-etal-2022-d4} as it offers diagnostically grounded assessment dialogues with explicit symptom-level annotations, making it well-suited for constructing structured clinical profiles.
The dataset contains 1{,}340 depression-related diagnostic interview sessions, each annotated with basic demographic information and mental disorder–related risk indicators.
In addition, D4 defines a structured psychiatric symptom framework with 41 symptom attributes, where each symptom is labeled as \emph{positive}, \emph{negative}, or \emph{not mentioned}.

\paragraph{Counseling Dialogue Data}
is primarily used to supplement the patient’s personality traits and speaking style.
We leverage the \emph{Client Reaction} annotations~\citep{li-etal-2023-understanding} as they capture patients’ affective responses and interaction patterns during counseling, providing valuable signals for modeling conversational style and personality-related behaviors.
The subset used in our study includes counseling dialogue histories from 300 clients.
However, the original dataset does not provide explicit annotations for basic demographic attributes or standardized symptom labels.

% \paragraph{Twitter-STMHD is used as the social media data source.}
%     It includes 6{,}803 users and approximately 37{,}000 posts associated with depression-related individuals.
%     Its provided labels are coarse-grained, mainly including a \emph{disorder flag} indicating whether a user is related to a mental health condition.
%     Nevertheless, its rich longitudinal posting histories enable the reconstruction of a patient's real-life experiences and symptom expressions over time.
%     At the same time, due to the large volume of posts, the dataset contains substantial irrelevant or low-quality content, which necessitates additional cleaning and filtering for our use case.
\paragraph{Social Media Data.}
% We use the Twitter-STMHD dataset~\citep{suhavi-etal-2022-twitter-stmhd} as it provides large-scale, longitudinal social media histories suitable for modeling real-world life events and symptom expressions over time.
% The dataset contains 6{,}803 users with approximately 37{,}000 posts related to depression.
% While the raw data include noisy and low-quality content, the longitudinal posting histories enable the reconstruction of temporally grounded patient experiences.
% To ensure sufficient temporal coverage, we focus on the depression cohort and retain only users with more than 500 tweets posted. \MY{suggested: To mitigate the inherent biases of social media data, such as performative posting or retrospective bias, we apply a multi-stage calibration pipeline. This includes: (1) filtering for high-engagement users (>500 tweets) to ensure sufficient longitudinal depth; (2) employing an LLM-based filter to exclude non-authentic, third-person, or retrospective narratives, retaining only in-situ personal experiences ; and (3) utilizing supervised classifiers specifically trained on clinical taxonomies to map noisy social expressions to standardized symptom vectors. This process transforms raw social streams into calibrated behavioral evidence.}
% \paragraph{Social Media Data.}
We utilize the Twitter-STMHD dataset \citep{suhavi-etal-2022-twitter-stmhd}, comprising 6,803 users and approximately 37,000 depression-related posts. To mitigate inherent biases such as performative posting or retrospective bias, we implement a multi-stage calibration pipeline: 
(1) \textbf{Longitudinal Filtering:} retaining high-engagement users ($>500$ tweets) to ensure temporal depth; 
(2) \textbf{Authenticity Verification:} employing an LLM-based filter to exclude non-authentic or third-person narratives, capturing only \textit{in-situ} personal experiences; and 
(3) \textbf{Clinical Mapping:} utilizing supervised classifiers to map noisy social expressions into standardized symptom vectors. This process transforms raw social streams into calibrated behavioral evidence.

%In the next section, we explain how missing information is supplemented via additional annotation and information extraction.

% Compared to existing datasets, this framework extends patient representation from a single-dimensional view to a three-dimensional, multi-faceted profile, allowing for more flexible and expressive prompt construction. Compared to the profile frameworks commonly adopted in existing patient simulation agents, our approach is grounded in real diagnostic dialogues and social media histories, employs more structured and clinically informed symptom labels, and incorporates substantially richer life event information.

\subsection{Annotation and Information Extraction}
In this section, we detail the preprocessing protocols for our three data sources to derive the standardized clinical and behavioral markers necessary for unified profile construction.

\paragraph{Basic Demographic Profile}
We established a unified demographic profile using a consistent annotation protocol.
For counseling dialogues, we achieved >95\% accuracy via a hybrid strategy: automatic labeling (Qwen3-8B) followed by manual verification and prompt refinement.
To scale annotation to social media posts, we first applied rule-based keyword filtering to select relevant posts, then annotated them using the refined prompt.
This process ensures reliable and consistent demographic labels across all three datasets.

% For counseling dialogue data, we adopt a hybrid annotation strategy that combines automatic labeling using Qwen3-8B, with manual verification and correction.
% We design a prompting scheme that achieves an annotation accuracy of over 95\%, and the full annotation prompt is provided in Appendix~\ref{promptappendixa}.
% For social media data, given the large volume of posts and the potential interference caused by users discussing others rather than themselves, we first apply rule-based filtering to select posts containing keywords related to demographic attributes.
% These filtered posts are then annotated by a large language model.
% Through this process, we obtain unified basic demographic profile across the three datasets.

\paragraph{Personality Traits}
We standardize personality representation using the widely adopted Big Five (OCEAN) model~\cite{jiang2024big5chat,azam2024llms}, ensuring consistency across data sources. To extract these traits from patient expressions, we employ a specialized LLM-based prompt (Appendix~\ref{promptappendixa}) that analyzes social media posts and counseling dialogues, outputting a 1–7 score for each dimension: Openness, Conscientiousness, Extraversion, Agreeableness, and Neuroticism.
% This design builds on recent work that infers Big Five traits from textual data using LLMs with embedding-based representations and chain-of-thought prompting.
% In our framework, we systematically apply this approach to multi-source depression-related data to obtain a unified and comparable personality representation at the patient level.

\paragraph{Symptom Attributes} 
To harmonize symptom descriptions across datasets, we aligned the 41 symptoms
% \MY{this needs explaination, citation and info in appendix} 
framework from the assessment dialogues provided by D4 dataset with the assessment dialogue data. Initial tests showed that zero‑shot LLM prompting performed poorly for symptom extraction. We therefore conducted supervised fine‑tuning with LoRA (rank $r=8$, scale $\alpha=32$) on the Qwen3‑8B backbone. After 10 epochs, the resulting classifier achieved >80\% average accuracy and >85\% precision. This optimized model was then applied to score 300 distinct client‑reaction segments, ensuring consistent symptom labeling.

% It is important to note that due to the inherent linguistic and structural disparities between social media posts and dialogue data, we adopted a distinct set of symptom tags for the social media corpus during the initial annotation phase. These specific definitions are detailed in the subsequent section.

\paragraph{Timeline Extraction} 
We utilize established supervised classifiers to extract, for each tweet, an 11‑dimensional life‑event score vector~\citet{Chen2024MappingLC} and a 38‑dimensional symptom relevance vector~\citet{Zhang2022SymptomIF}. High‑confidence signals are retained via a hard threshold $p$.

To ensure the timeline reflects the user’s in‑situ experience, we first apply an LLM‑based filter is used to discard posts containing retrospective descriptions or third‑person narratives (e.g., discussing a friend's diagnosis). Only posts describing current, first‑person sensations or events are retained. This process results in a high‑fidelity, chronological sequence of symptom markers and life events for each user.

% Leveraging the user-level annotations provided by the Twitter-STMHD dataset, we anchor each user's timeline to their self-reported diagnosis date. We designate the timestamp of the diagnosis disclosure as the reference point $t_{\text{diag}}$. For every post $i$ with a timestamp $t^{(i)}$, we compute a relative timestamp $T_{\text{ref}}^{(i)} = t^{(i)} - t_{\text{diag}}$ (in days). This alignment allows us to explicitly differentiate between prodromal signals ($T_{\text{ref}} < 0$) and post-diagnosis behavioral changes ($T_{\text{ref}} \geq 0$), providing a temporal context critical for analyzing disease progression.

%To ensure the identified symptoms and events reflect the user's \textit{in-situ} experiences, we employ an LLM-based filter. The model is prompted to discern the temporal currency and experience attribution of each post. Only posts where the user describes a current sensation or a first-person life event are retained; retrospective memories or general commentary are discarded. This rigorous filtering yields a high-fidelity sequence of valid symptom markers and life events for each user.

% \begin{figure}[htbp]
%     \centering
%     \includegraphics[width=0.9\columnwidth]{latex/figures/paired.png}
%     \caption{Two-stage label alignment across data sources.}
%     \label{fig:paireddata}
% \end{figure}
% \vspace{-4mm}
\subsection{Profile Attribute Alignment}
We implement a two‑stage matching process to construct unified, multi‑source patient profiles, as illustrated in Figure~\ref{fig:deprofile_full}.
% \MY{you mention C-A profile too many times without explaination, i change it into skeleton profile to match the figure content. you better have a number i.e. how many profiles with how many attributes/components, and mention this is by far the largest profile library with most components in the intro contribution part}
% A core challenge in our framework is that different data sources provide complementary yet heterogeneous user information. To address this, we integrated the classified and annotated datasets (Assessment, Counseling, and Social Media) through a two-stage matching process to construct unified patient profiles.

\paragraph{Stage 1: Deterministic Matching}
We first create a coherent clinical “skeleton” by aligning Counseling and Assessment dialogues. For each matched pair, we unify the Basic Demographic Profile, Disorder Risk, Personality Traits, and Symptom Attributes. To ensure clinical consistency, we enforce hard matching constraints: a match is valid only if all demographic tags align and no symptom labels conflict (e.g., a profile cannot simultaneously possess and lack the same symptom). This rigorous filtering yields 3,258 conflict‑free profiles, forming a reliable foundation for further expansion.

\paragraph{Stage 2: Similarity-Driven Candidate Expansion.}
Building upon the verified \textbf{skeleton}, this stage aims to enrich the static ``skeleton'' into a multifaceted persona by retrieving aligned profiles.
% We employ a hybrid matching strategy combining hard demographic constraints with soft attribute similarity to ensure both \textit{clinical consistency} and \textit{persona fidelity}.
Starting from each skeleton, we retrieve aligned social‑media profiles to build a richer, longitudinal persona. 1) Demographic pre‑filtering. We first apply hard matching on the Basic Demographic Profile to obtain an initial candidate pool; 2)~Symptom mapping. To bridge different symptom taxonomies, we project all symptom tags into a shared space of 20 clinically‑grounded categories and prune candidates that exhibit clinical conflicts. The complete list of matched symptom attributes across data sources
is provided in Appendix~\ref{app:appendix_symptoms}; 3)~Similarity scoring. For each remaining candidate, we compute two alignment metrics:
\begin{itemize}[leftmargin=*,noitemsep,topsep=0pt]
    \item \textbf{Symptom Consistency Score}:
    \[
    \text{Sim}_{\text{symp}} = \frac{|S_{C\text{-}A}^{+} \cap S_{\text{Social}}|}{|S_{\text{Social}}|},
    \]
    where $S_{C\text{-}A}^{+}$ denotes the positive symptom set of the C–A profile and $S_{\text{Social}}$ denotes the symptom set of the social media candidate. This metric penalizes candidates that introduce extraneous symptoms.
    
    \item \textbf{Personality Similarity}: Cosine similarity between Big Five trait vectors ($\text{Sim}_{\text{pers}}$).
\end{itemize}

4) High-precision selection. We further retain only candidates meeting strict thresholds: $\text{Sim}_{\text{symp}} > 0.8$ and $\text{Sim}_{\text{pers}} > 0.8$. Qualified candidates are ranked by the composite score:
\[
\text{Sim}_{\text{total}} = \text{Sim}_{\text{symp}} + \text{Sim}_{\text{pers}}.
\]

While the alignment between clinical skeletons and social media timelines relies on heuristic thresholds, these act as high-precision gatekeepers rather than loose associations. The $Sim > 0.8$ threshold for both symptoms and personality ensures that retrieved timelines match the core psychopathological and dispositional structure, rather than merely resembling the clinical profile. This design assumes that although individuals differ, the causal and temporal relationships between life events and symptom dynamics are generalizable across patients sharing similar demographic and clinical skeletons. By leveraging these shared patterns, \textsc{DEPROFILE} shifts from \textit{individual-level} matching to \textit{pattern-level} clinical grounding.

This protocol yields an average of \textbf{88 high-quality social media profiles per skeleton}, enriching longitudinal context while maintaining clinical coherence. To validate the alignment, we conducted a manual review of 30 randomly sampled profiles, where six clinical researchers examined potential internal contradictions. The results show that the constructed profiles are \textbf{highly coherent}, with no fundamental conflicts between timeline events and clinical symptoms. This suggests that the combination of demographic constraints and high-similarity thresholds effectively preserves \textit{temporal and clinical authenticity}. A full example is provided in Appendix~\ref{app:deprofile_example}.

\section{Patient Simulation Framework}
After obtaining 3,258 unified patient profiles, each associated with an average of 88 candidate timelines, our goal is to organize this information for realistic patient simulation. We therefore structure the profiles into prompts that support dynamic and temporally grounded behaviors, transforming static attributes and longitudinal records into clinically coherent conversational agents. The following subsections present our three-component solution: a memory structuring agent, a modular prompting strategy, and a standardized evaluation protocol.
% \MY{you need something here. not in this section we talk about xxx but the motivation, i.e. after obtaining xxx profile and corresponding data, we xxxx}
\subsection{The Chain-of-Change (CoC) Agent}
\label{sec:coc_agent}
\begin{figure}
    \centering
    \includegraphics[width=\columnwidth]{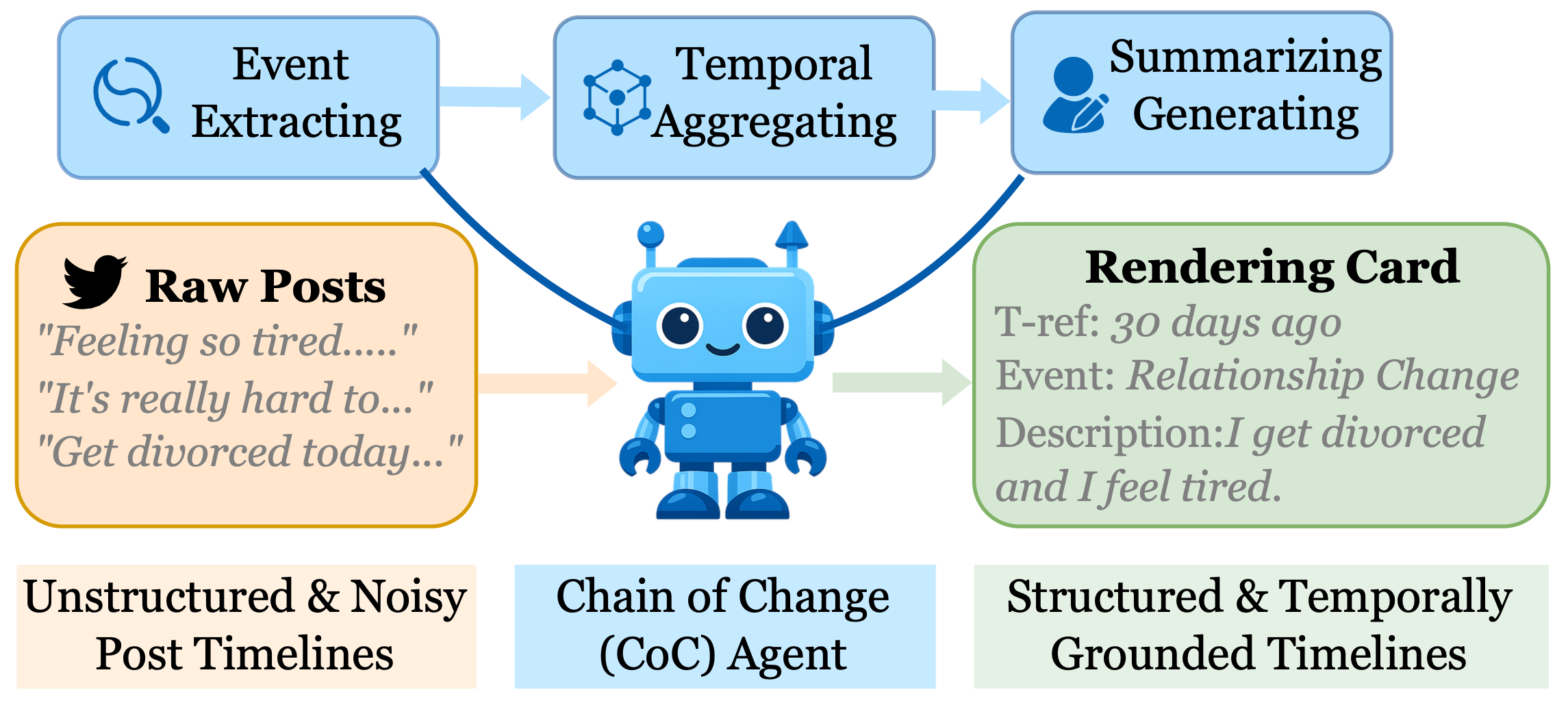}
    \caption{Illustrative workflow of the LLM-based Chain of Change (CoC) Agent.}
    \label{fig:cocagent}
\vspace{-1.5em}
\end{figure}

To convert a raw longitudinal timeline of social media posts into a structured, retrievable memory system, we introduce the \textbf{Chain-of-Change (CoC) Agent}. This agent operates through three sequential phases, as illustrated in Figure~\ref{fig:cocagent}.
% To bridge the gap between raw, noisy social media streams and the rigorous temporal consistency required for clinical simulation, we introduce the \textbf{Chain-of-Change (CoC) Agent}. 
% Unlike previous approaches that feed raw posts or simple labels directly into the prompt, the CoC Agent transforms the longitudinal timeline into a \textit{retrievable, verifiable, and structured memory system}.The pipeline operates in three phases illustrated in Figure~\ref{fig:cocagent}:

\paragraph{Semantic Structuring and Node-identification.}
Given a simulation anchor day $t_{\text{anchor}}$ and a look-back horizon $H$, the agent first retrieves relevant historical posts. Each post is then structured into semantic triplets accompanied by a natural-language abstract: For \textbf{symptom-related posts}, \textit{Subject, Experience} and \textit{Severity/Frequency} are extracted. For \textbf{life-event posts},\textit{Actor, Action} and \textit{Impact} are listed. 
This representation preserves expressive nuance while enabling consistent downstream processing.

% To preserve the nuance of user expression while enabling structural alignment, we move beyond simple label matching:
% \begin{itemize}
%     \item \textbf{Symptom Normalization:} For each symptom record, while retaining the canonical label for alignment, the agent employs an instruction-tuned extractor to generate a \textit{Symptom Triplet} (Subject, Experience, Severity/Frequency) and a natural language abstract. This prevents the loss of specific physiological descriptions (e.g., ``head feels like splitting'' vs. generic ``Headache'').
%     \item \textbf{Event Abstraction:} Similarly, raw life event posts are processed into \textit{Event Triplets} (Actor, Action, Impact) to filter out irrelevant daily noise and retain only clinically significant narrative beats.
% \end{itemize}

% \paragraph{2. Temporal Aggregation and Relation Building.}
% Individual nodes are then organized into a directed temporal graph. The agent explicitly models two types of edges: \textit{Temporal Precedence} (sequential ordering of events) and \textit{Symptom Persistence} (linking recurring symptoms to track chronicity).
% To manage context length and simulate human episodic memory, these nodes are aggregated into \textbf{Episodes} using a fixed sliding window (e.g., 7 days). Each episode $\mathcal{E}_i$ encapsulates the progression of symptoms and events within that distinct timeframe.

\paragraph{Temporal Aggregation and Relation Building.}
Individual nodes are organized into a directed temporal graph with two edge types:
\begin{itemize}[leftmargin=*,noitemsep,topsep=0pt]
\item \textit{Temporal Precedence}: captures event ordering.
\item \textit{Symptom Persistence}: links recurring symptoms across time.
\end{itemize}
To manage context length, nodes are aggregated into \textbf{Episodes} using a fixed sliding window (e.g., 7 days). Each episode $\mathcal{E}_i$ summarizes the symptoms and events within its time span.

\paragraph{Memory Rendering and Reference Protocol.}
Each episode is rendered into a human-readable \textbf{Memory Card}, containing a concise summary and a \textbf{Relative Timestamp} $T_{\text{ref}}$ (e.g., ``3 days ago'' relative to $t_{\text{anchor}}$). We specifically design a dual-channel output mechanism:
\begin{itemize}[leftmargin=*,noitemsep,topsep=0pt]
    \item \textbf{Knowledge Graph:} used for automated evaluation to verify factual alignment between generated responses and ground-truth events.
    \item \textbf{Rendered Memory Cards:} injected into the simulation prompt under a \textit{Strict Citation Protocol}, which requires explicit reference to $T_{\text{ref}}$ when discussing past events and prohibits fabrication of unspecified details.
\end{itemize}

This design ensures narrative traceability while maintaining conversational naturalness.

% \paragraph{3. Memory Rendering and Reference Protocol.}
% To maximize the downstream LLM's adherence to the timeline, the CoC Agent renders the structured episodes into human-readable \textbf{``Memory Cards.''}
% Each card contains a dense natural language summary of the episode and, crucially, establishes a \textbf{Relative Timestamp} $T_{\text{ref}}$ (e.g., ``3 days ago'' relative to $t_{\text{anchor}}$).

% \textbf{Dual-Channel Output.} 
% The CoC Agent ultimately outputs two distinct data structures:
% \begin{itemize}
%     \item \textbf{The Knowledge Graph:} Used for automated evaluation to strictly verify if the generated response aligns with ground-truth facts (Alignment Check).
%     \item \textbf{The Rendered Cards:} Injected into the patient simulation prompt. We enforce a \textit{Strict Citation Protocol}: the model is constrained to explicitly cite the $T_{\text{ref}}$ from the cards when discussing past events and is strictly prohibited from fabricating specific dates or details not present in the cards.
% \end{itemize}
% This design ensures that the simulated patient's narrative remains traceable to the source data while maintaining the conversational naturalness of a real user.

\subsection{Modular Prompt Construction}

We hereby condition the LLM to role-play a patient through a modular system prompt, assembled from the following components (detailed in Appendix~\ref{app:prompt-components}):

\begin{itemize}[leftmargin=*,noitemsep,topsep=0pt]
\item \textbf{Basic Profile \& Risk:} Basic demographics are coupled with a Disorder Risk statement specifying BDI-II severity levels and explicit suicide-risk signals for safety conditioning, following \citet{cohen2025evaluating}.
\item \textbf{Personality Traits:} OCEAN scores are discretized into three levels (Low/Medium/High) using thresholds \( \leq 2 \), \( (2,5] \), and \( > 5 \). These are expressed as natural‑language instructions rather than raw scores, inspired by \citet{dai2025profile}.

\item \textbf{Symptom Attributes:} Each symptom label is rewritten into a natural‑language description (with paired positive/negative realizations). Attributes are further enriched with temporally ordered symptom expressions derived from the CoC Agent’s graph, enabling time‑aware reporting without hallucination.

\item \textbf{Dialogue Demonstrations:} Few‑shot exemplars are selected from \textbf{Assessment} data to model clinical interview dynamics and from \textbf{Counseling} data to capture therapeutic client reactions.

\item \textbf{Life Event Timeline:} A sequence of event cards derived from the CoC Agent grounds the narrative in a specific, event‑driven history, ensuring temporal coherence.
\end{itemize}
% Component \textbf{S} provides a structured checklist of positive and negative symptom attributes derived from the D4 diagnostic framework. Rather than exposing raw attribute labels, we rewrite each attribute into clinically grounded natural-language descriptions (with paired positive vs.\ negative realizations) so that the model can answer conversationally while preserving clinical consistency. When available, symptom attributes are further enriched with temporally ordered symptom expressions (e.g., relative time points) derived from STMHD-based symptom timelines, enabling time-aware symptom reporting without hallucinating unsupported details.
 
% \noindent\textbf{Dialogue Demonstrations:} Serves as few-shot style exemplars. We select \textbf{Assessment} data to model clinical interview dynamics and \textbf{Counseling} data to capture therapeutic client reactions.
 
 %\noindent\textbf{Life Event Timeline:} Ensures temporal coherence. A sequence of event cards derived from Twitter-STMHD by CoC Agent is injected to ground the narrative in a specific, event-driven history.

\subsection{Question Evaluation Protocol }
To evaluate profile‑grounded patient simulation in a structured and reproducible manner, we design a three‑stage question protocol (full list in Appendix~\ref{appendixB}):

\paragraph{Stage 1: Persona Profiling QA.}
This stage elicits basic demographic information and personality traits. For personality, we use open‑ended questions inspired by \citet{wang-etal-2024-incharacter} to evaluate the simulation’s capacity to reflect the OCEAN profile. This stage primarily assesses \emph{persona faithfulness} and stylistic stability.

\paragraph{Stage 2: Symptom Attribute QA.}
Questions are formulated based on symptom attributes distilled from assessment‑dialogue data, probing standardized clinical attributes in a conversational setting. This stage evaluates \emph{symptom consistency} with the structured profile and robustness under safety‑critical prompts.

% \paragraph{Stage 3: Life‑event Timeline QA.}
% A phased timeline interview probes event‑driven state evolution:
% \begin{enumerate}[leftmargin=*,noitemsep,topsep=0pt]
% \item A salient event and the immediate reaction/impact.
% \item An additional event and its downstream influence on mood/functioning.
% \item The current overall status, recent changes, and anticipated future impact.
% \end{enumerate}
% This stage evaluates whether the model can produce coherent, temporally grounded narratives that reflect plausible dynamic trajectories.

\paragraph{Stage 3: Life-event Timeline QA.}

A phased timeline interview probes event-driven state evolution, including (1) a salient event and its immediate impact, (2) an additional event and its downstream influence on mood/functioning, and (3) the current status, recent changes, and anticipated future trajectory. This stage evaluates whether the model can produce coherent, temporally grounded narratives reflecting plausible dynamic progression.

% =========================
\section{Experimental Design}
To thoroughly evaluate our patient simulation framework, we design a comprehensive evaluation protocol. Our primary goal is to quantify \textit{how different profile components contribute} to simulation quality—an analysis not previously conducted in detail. We first analyze the contribution of individual attributes within our framework and then benchmark the complete system against state-of-the-art methods.

\subsection{Comparison Methods}
\paragraph{Systematic Evaluation of Profile Attributes}
We conduct a controlled evaluation to measure the impact of each structured profile component. Starting from a minimal foundation, we incrementally add attributes to assess their individual and combined effects on simulation quality.

\begin{itemize}[leftmargin=*,noitemsep,topsep=0pt]
\item \textbf{Basic Profile} : Serves as the foundation, containing only core demographics (B), disorder risk level (R), and personality traits (P).
\item \textbf{+ Timeline (basic+T)}: Adds the chronological Life Event Timeline (T) to evaluate the isolated effect of temporal grounding.
\item \textbf{+ Symptoms (basic+S)}: Incorporates the structured clinical symptom checklist (S) to measure the contribution of explicit psychopathological attributes.
\item \textbf{+ Symptoms \& Timeline (basic+ST)}: Combines symptoms (S) with the timeline (T) to test their synergistic effect on clinical narrative coherence.
\item \textbf{Full Framework without Timeline (full w/o T)}:
% \MY{full and Deprofile are really easy to confuse. name a different way? why don't you name full as Deprofile as they have better results. Current Deprofile can be something minus temporal} 
A strong, high-resource baseline including few-shot dialogue demonstrations but excluding the timeline. Comparing this to our complete framework (\textsc{Deprofile}) isolates the specific value of factual temporal constraints when other contextual signals are present.
\end{itemize}

\paragraph{Benchmarking Against Prior Work}
We compare our evidence-grounded profiles against two recent state-of-the-art mental health disorder patient simulation systems that employ explicit, yet synthetically generated, profile specifications. Their profile structures are aligned with our prompt format for a controlled comparison.
\begin{itemize}[leftmargin=*,noitemsep,topsep=0pt]
\item \textsc{Patient-$\Psi$}~\cite{wang2024patient}: We utilize all 106 profiles from the official release.
\item \textsc{Eeyore}~\cite{liu2025eeyore}: We perform stratified sampling based on key demographic and clinical attributes to select 200 representative profiles.
\end{itemize}
This comparison assesses the advantage of data-driven profiles over synthetic specifications in achieving clinical fidelity and behavioral realism.

\subsection{Evaluation Metrics}
We employ automated embedding-based metrics and a targeted LLM-as-a-Judge protocol to evaluate performance across multiple dimensions.
% 【TODO】
% 1. 生成质量指标（BLEU / ROUGE / DIST）
% 2. 症状一致性与覆盖率
% 3. 时间一致性（timeline）
% 4. 诊断相关评估（分类 / 医生判断）
\paragraph{MentalBERT-based Metrics for Realism and Diversity}
Using Chinese MentalBERT~\citep{zhai2024chinesementalbert}, we compute \textbf{Realism} (cosine similarity between generated and real patient utterance embeddings) and \textbf{Inter-Patient Diversity} (Q-Centroid). Let $\mathbf{e}(x) \in \mathbb{R}^d$ denote the embedding of $x$. For patient $i$ with responses $\mathcal{G}_i$ and real utterances $\mathcal{R}_i$:
\begin{equation}
\bar{\mathbf{g}}_i=\frac{1}{T_i}\sum_{t=1}^{T_i}\mathbf{e}(g_{it}),\quad
\bar{\mathbf{r}}_i=\frac{1}{M_i}\sum_{m=1}^{M_i}\mathbf{e}(r_{im}),
\end{equation}
and define Realism as:
\begin{equation}
\mathrm{Realism}=\frac{1}{N}\sum_{i=1}^{N}\cos\!\left(\bar{\mathbf{g}}_i,\bar{\mathbf{r}}_i\right).
\end{equation}
For each question $q$, let $\mathbf{c}_q = \frac{1}{N_q} \sum_{j=1}^{N_q} \mathbf{z}_{j,q}$ be the centroid of answers. The question-wise diversity is
\begin{equation}
\mathrm{Div}_{\mathrm{inter}}(q)
=\frac{1}{N_q}\sum_{i=1}^{N_q}\left[1-\cos\!\left(\mathbf{z}_{i,q}, \mathbf{c}_q\right)\right],
\end{equation}
and the overall diversity is averaged over questions:
\begin{equation}
\mathrm{Div}_{\mathrm{inter}}=\frac{1}{|\mathcal{Q}|}\sum_{q\in\mathcal{Q}}\mathrm{Div}_{\mathrm{inter}}(q).
\end{equation}

\paragraph{LLM-as-a-Judge (G-Eval)}
Following G-Eval~\citep{zhang2023geval}, we use GPT-5 to score (1--5) each response on persona faithfulness, event richness, and symptom consistency, given the full patient prompt and dialogue. The full G-Eval
 prompt is provided in Appendix~\ref{app:g-eval}.

\paragraph{Human Evaluation}
To validate automated metrics, we conduct a human evaluation. We recruit eight psyciatric experts. From our full profile set, we select 30 representative patient profiles (stratified by demographics and G-Eval scores). Each profile is evaluated by exactly three reviewers, who rate each dialogue on a 1--5 Likert scale for overall quality. Reviewers are blind to model identity. We compare five systems: basic, basic+S, \textsc{Patient-$\Psi$}, \textsc{Eeyore}, and \textsc{Deprofile}. In total, 450 ratings are collected (30 profiles $\times$ 5 models $\times$ 3 raters). The human evaluation guidelines and recruitment details are provided in Appendix~\ref{app:human-eval}.

% =========================
\section{Results and Analysis}
\label{sec:results}
% Following our experimental design, we present a structured analysis of the results. We first quantify the contribution of each profile component through controlled ablations, then benchmark our complete framework against state-of-the-art methods, and finally examine the unique role of the temporal module and the framework's robustness.
% \subsection{Contribution of Profile Components}
% Table~\ref{tab:mentalbert_results} and Table~\ref{tab:semantic_results} systematically evaluates the performance of incremental variants of our framework, from a minimal basic profile to the full \textsc{Deprofile} system. The standard deviations and 95\% confidence intervals are reported in Table~\ref{tab:ablation_ci_geval} in Appendix~\ref{app:fullresults}.

% \label{sec:results}
Following our experimental design, we present a structured analysis: first quantifying the contribution of each profile component via ablations, then benchmarking against state-of-the-art methods, and finally examining the role of the temporal module and overall robustness.
\subsection{Contribution of Profile Components}
Table~\ref{tab:ablation_metrics} systematically evaluates the performance of incremental variants of our framework, from a minimal basic profile to the full \textsc{Deprofile} system. The standard deviations and 95\% confidence intervals are reported in Table~\ref{tab:ablation_ci_geval} in Appendix~\ref{app:fullresults}.

\paragraph{The Dual Role of the Timeline Module.}
The timeline exhibits a context-dependent function, acting as either an \textit{enricher} or a \textit{constraint}.
\begin{itemize}[leftmargin=*,noitemsep,topsep=0pt]
\item \textbf{Enrichment in Sparse Contexts:} When the profile is minimal (e.g., \textit{Basic Profile}), adding the timeline significantly increases response diversity and semantic similarity. Here, the timeline provides essential narrative details, enriching an otherwise flat persona.
\item \textbf{Constraint in Rich Contexts:} In the full-context setting with few-shot demonstrations, adding the timeline to create the complete \textsc{Deprofile} causes a slight decrease in similarity  and diversity. This reflects a shift from ungrounded fluency to factual precision. The dialogue-only variant (\textit{w/o Timeline}) tends to hallucinate diverse but unverified details. The timeline curbs this, trading hallucinated variety for verifiable accuracy, as confirmed by a substantial gain in G-Eval \textit{Event Richness} ($4.57 \to 4.86$).
\end{itemize}

% \begin{table}[htbp]
% \centering
% \small
% \setlength{\tabcolsep}{4pt}
% \caption{MentalBERT-based automatic evaluation across profile configurations.}
% \label{tab:mentalbert_results}
% \begin{tabular}{lcc}
% \toprule
% \textbf{Profile Setting} & \textbf{Similarity} $\uparrow$ & \textbf{Q-Centroid} $\uparrow$ \\
% \midrule
% basic        & 0.9312 & 0.0540 \\
% basic+T      & 0.9442 & 0.0587 \\
% \midrule
% basic+S       & 0.9360 & 0.0608 \\
% basic+ST     & 0.9407 & 0.0675 \\
% \midrule
% full w/o T      & \textbf{0.9414} & \textbf{0.0656} \\
% \textsc{Deprofile} & 0.9393 & 0.0628 \\
% \bottomrule
% \end{tabular}
% \end{table}
% \vspace{-3mm}
% \begin{table}[htbp]
% \centering
% \small
% \setlength{\tabcolsep}{4pt}
% \caption{G-Eval results (Persona Faithfulness, Event Richness and Symptom Consistency) for clinical narrative quality across different profile configurations.}
% \label{tab:semantic_results}
% \begin{tabular}{lccc}
% \toprule
% \textbf{Profile Setting} & \textbf{Persona} & \textbf{Event} & \textbf{Symptom} \\
% \midrule
% basic        & 4.71 & 4.60 & 3.99 \\
% basic+T & 4.63 & 4.84 & 3.89 \\
% \midrule
% basic+S & 4.94 & 4.80 & 4.84 \\
% basic+ST     & 4.80 & \textbf{4.90} & 4.71 \\
% \midrule
% full w/o T    & \textbf{4.97} & 4.57 & \textbf{4.86} \\
% \textsc{Deprofile} & 4.87 & 4.86 & 4.79 \\
% \bottomrule
% \end{tabular}
% \end{table}
\begin{table}[htbp]
\centering
\small
\setlength{\tabcolsep}{4pt}
\caption{Automatic evaluation across profile configurations. Sim=Similarity; Q-Cent=Q-Centroid; Pers=Persona Faithfulness; Evt=Event Richness; Sym=Symptom Consistency.}
\label{tab:ablation_metrics}
\begin{tabular}{lcc|ccc}
\toprule
\multirow{2}{*}{\textbf{Profile}} & \multicolumn{2}{c|}{\textbf{MentalBERT}} & \multicolumn{3}{c}{\textbf{G-Eval}} \\
\cmidrule(lr){2-3} \cmidrule(lr){4-6}
& \textbf{Sim} $\uparrow$ & \textbf{Q-Cent} $\uparrow$ & \textbf{Pers} & \textbf{Evt} & \textbf{Sym} \\
\midrule
basic        & 0.931 & 0.054 & 4.71 & 4.60 & 3.99 \\
basic+T      & 0.944 & 0.059 & 4.63 & 4.84 & 3.89 \\
basic+S      & 0.936 & 0.061 & 4.94 & 4.80 & 4.84 \\
basic+ST     & 0.941 & 0.068 & 4.80 & \textbf{4.90} & 4.71 \\
full w/o T   & \textbf{0.941} & \textbf{0.066} & \textbf{4.97} & 4.57 & \textbf{4.86} \\
\textsc{Deprofile} & 0.939 & 0.063 & 4.87 & 4.86 & 4.79 \\
\bottomrule
\end{tabular}
\end{table}

\FloatBarrier
\begin{table*}[!t]
\centering
% \scriptsize
\fontsize{8}{9}\selectfont
\setlength{\tabcolsep}{5pt}
\caption{Main results comparing \textsc{Deprofile} with \textsc{Patient-$\Psi$} (\textsc{P-$\Psi$}) and \textsc{Eeyore} across different LLM backbones.
Metrics include dialogue realism (MentalBERT similarity), response diversity (Q-Centroid), and event richness.}
\label{tab:main_results}
\begin{tabular}{l|ccc|ccc|ccc}
\toprule
\multirow{2}{*}{\textbf{Backbone Model}} 
& \multicolumn{3}{c|}{\textbf{Realism} $\uparrow$} 
& \multicolumn{3}{c|}{\textbf{Diversity(Q-centroid)} $\uparrow$} 
& \multicolumn{3}{c}{\textbf{Event Richness} $\uparrow$} \\
& \textsc{P-$\Psi$} & \textsc{Eeyore} & \textsc{Deprofile} 
& \textsc{P-$\Psi$} & \textsc{Eeyore} & \textsc{Deprofile} 
& \textsc{P-$\Psi$} & \textsc{Eeyore} & \textsc{Deprofile} \\
\midrule
\textit{Llama-3.1-8B-Instruct} 
& 0.888 & 0.896 & \textbf{0.936}
& 0.085 & 0.085 & \textbf{0.085}
& 3.74 & 3.99 & \textbf{4.24} \\

\textit{Qwen-3-4B-Instruct} 
& 0.893 & 0.898 & \textbf{0.913}
& 0.024 & 0.020 & \textbf{0.039}
& 4.76 & 4.83 & \textbf{4.84} \\

\textit{GPT-4o-mini} 
& 0.927 & 0.925 & \textbf{0.940}
& 0.026 & 0.029 & \textbf{0.050}
& 2.95 & 3.08 & \textbf{4.39} \\

\textit{DeepSeek-V3.2-Exp} 
& 0.923 & 0.922 & \textbf{0.939}
& 0.046 & 0.044 & \textbf{0.063}
& 4.76 & 3.77 & \textbf{4.85} \\
\midrule
\textit{Average} 
& 0.908 & 0.910 & \textbf{0.932}
& 0.045 & 0.044 & \textbf{0.059}
& 4.05 & 3.92 & \textbf{4.58} \\
\bottomrule
\end{tabular}
\end{table*}

\begin{table*}[t]
\centering
\fontsize{8}{9}\selectfont
\setlength{\tabcolsep}{5pt}
\caption{Human evaluation results. Mean $\pm$ std across profiles. Faith: Persona Faithfulness; Const: Symptom Consistency. Win Rate: proportion of profiles where \textsc{Deprofile} is rated higher than the row model. Adjacent\%: proportion of profile$\times$model pairs where the three raters' scores differ by at most 1 point. Wilcoxon signed-rank test (Deprofile vs.\ that baseline): $^{\dagger}$ $p < 0.001$, $^{*}$ $p < 0.05$. Significance shown only for baselines.}
\label{tab:human_eval}
\begin{tabular}{lccccccc}
\toprule
\textbf{Model} & \textbf{Realism} & \textbf{Persona Faith} & \textbf{Event Richness} & \textbf{Symptom Const} & \textbf{Overall} & \textbf{Win Rate} & \textbf{Adjacent\%} \\
\midrule
basic        & 3.02 $\pm$ 0.48$^{\dagger}$ & 3.55 $\pm$ 0.60 & 2.55 $\pm$ 0.63$^{\dagger}$ & 3.35 $\pm$ 0.77$^{*}$ & 3.09 $\pm$ 0.45$^{\dagger}$ & 80.0\% & 60.0\% \\
basic+S      & 3.20 $\pm$ 0.62$^{*}$ & \textbf{3.90 $\pm$ 0.77} & 2.82 $\pm$ 0.66$^{*}$ & 3.75 $\pm$ 0.85 & 3.39 $\pm$ 0.37$^{*}$ & 63.3\% & 43.3\% \\
\textsc{P-$\Psi$} & 2.22 $\pm$ 0.81$^{\dagger}$ & 2.54 $\pm$ 0.58$^{\dagger}$ &  3.33 $\pm$ 1.04$^{*}$ & 2.68 $\pm$ 0.97$^{\dagger}$ & 2.73 $\pm$ 0.55$^{\dagger}$ & 86.7\% & 30.0\% \\
\textsc{Eeyore}   & 2.16 $\pm$ 0.94$^{\dagger}$ & 2.67 $\pm$ 0.93$^{\dagger}$ & 3.06 $\pm$ 0.86$^{*}$ & 2.27 $\pm$ 0.90$^{\dagger}$ & 2.58 $\pm$ 0.62$^{\dagger}$ & \textbf{90.0\%} & 46.7\% \\
\textsc{Deprofile} & \textbf{3.67 $\pm$ 0.59} & 3.75 $\pm$ 0.75 & \textbf{3.49 $\pm$ 0.85} & \textbf{3.86 $\pm$ 0.65} & \textbf{3.73 $\pm$ 0.41} & --- & \textbf{76.7\%} \\
\bottomrule
\end{tabular}
\end{table*}

\paragraph{Essential Role of Symptom Attributes.}
The structured symptom checklist ($S$) is critical for clinical validity. Variants lacking explicit symptom attributes (e.g., \textit{Basic Profile}) score poorly on symptom consistency ($<4.0$). Introducing the checklist (in the \textit{+Symptoms} variant) sharply corrects this, achieving a score of $4.84$. This demonstrates that explicit symptom grounding is necessary for the model to accurately adhere to a clinical profile.

\subsection{Benchmarking Against Prior Work}
Table~\ref{tab:main_results} compares \textsc{Deprofile} with two representative prior systems, \textsc{Patient-$\Psi$} and \textsc{Eeyore}, across four LLM backbones. The standard deviations and 95\% confidence intervals are reported in Table~\ref{tab:appendix_stats_transposed} in Appendix~\ref{app:fullresults}.

\paragraph{Superior Realism and Fidelity.}
\textsc{Deprofile} consistently achieves higher semantic realism (MentalBERT Similarity) across all models. On smaller open-source models like \textit{Llama-3.1-8B}, it delivers a substantial improvement ($+4.8\%$ over \textsc{Patient-$\Psi$}), indicating that our evidence-driven profiles effectively guide the LLM to mimic the linguistic patterns of real patients more accurately than synthetic or seed-based approaches.

\paragraph{Enhanced Behavioral Diversity.}
A key limitation of prior methods is their tendency toward generic, homogeneous responses. \textsc{Deprofile} demonstrates significantly higher inter-patient diversity (Question Centroid metric). For instance, using \textit{GPT-4o-mini} and \textit{DeepSeek-V3.2}, our method nearly doubles the diversity score compared to \textsc{Eeyore} ($0.050$ vs. $0.029$; $0.063$ vs. $0.044$). This shows that incorporating longitudinal and symptom-specific data preserves the unique characteristics of individual profiles, mitigating the ``averaging effect'' common in LLM simulation.

\subsection{Human Evaluation}

The human evaluation results in Table~\ref{tab:human_eval} are consistent with trends observed in automatic metrics. Both experts and LLM-based judges (G-Eval) indicate that incorporating explicit symptom checklists effectively anchors clinical fidelity. For example, the \texttt{basic+S} variant achieves a slightly higher Symptom Consistency score than \textsc{Deprofile} (3.75 vs.\ 3.56), suggesting that dense symptom prompts help constrain hallucinations.
Beyond this, \textsc{Deprofile} demonstrates superior overall clinical plausibility, significantly outperforming all baselines (Wilcoxon signed-rank test, $p < 0.05$). It achieves the highest mean scores across key dimensions, particularly in Realism (3.67) and Overall Quality (3.73 $\pm$ 0.41). In pairwise comparisons, \textsc{Deprofile} attains win/tie rates of 63.3\%--90.0\% across profiles, with the largest margin over \textsc{Eeyore}. Additionally, it shows the highest inter-rater consistency (76.7\% Adjacent\%), indicating more coherent and unambiguous clinical evidence for expert assessment.

% =========================
%\section{Discussion}

\subsection{Temporal Mechanism and Factualness}
Our study
% \MY{unfinished add specific results comparison here} 
confirms the timeline module's dual regulatory mechanism: it enriches sparse profiles and constrains rich ones toward factual consistency. Furthermore, \textsc{Deprofile} exhibits strong cross-model robustness. Even on smaller open-source models (e.g., Qwen-3-4B-Instruct, Llama-3.1-8B), it achieves clinical fidelity comparable to that of closed-source large models. This finding suggests that for medical simulation, constructing high-quality, structured longitudinal data can be more effective for combating ``hollowness'' and hallucination than merely scaling model parameters.

\section{Related Work}

\paragraph{Profile Modeling Frameworks}
Recent frameworks for depression-oriented patient simulation include \textsc{Patient-$\Psi$}~\cite{wang2024patient}, Roleplay-doh~\cite{louie2024roleplay}, AnnaAgent~\cite{wang2025annaagent}, and \textsc{Eeyore}~\cite{liu2025eeyore}. These approaches typically rely on LLMs to expand a limited set of symptoms, stressors, or personality tags into narrative patient profiles that drive role-play interactions. While often fluent and psychologically plausible, such profiles are largely model-fabricated rather than grounded in real, verifiable longitudinal data. As a result, the structured information encoded in these systems is usually restricted to a small number of labels (e.g., severity or traits), with important factors such as life-event sequences, social and occupational trajectories, and comorbid symptoms rarely modeled in a systematic or retrievable manner~\cite{louie2024roleplay,wang2025annaagent}. Consequently, reasoning about emotions or causal mechanisms in systems like \textsc{Patient-$\Psi$} and \textsc{Eeyore} relies heavily on the model’s internal inference rather than evidence anchored in real-world timelines~\cite{wang2024patient,liu2025eeyore}, limiting external verifiability and potentially amplifying model biases.

\paragraph{Mental Health Datasets}
A variety of datasets have been developed for mental health NLP, each capturing complementary aspects of patient behavior.
D4 provides clinically grounded Chinese diagnostic dialogues for depression, supporting symptom-level modeling and severity-related tasks~\citep{yao-etal-2022-d4}.
In counseling settings, \citep{li-etal-2023-understanding} introduce a theoretically grounded annotation framework to study client reactions to counselor strategies.
Complementary to dialogue-based resources, Twitter-STMHD offers large-scale, user-level social media data with temporal context for multiple mental health disorders, enabling longitudinal analyses~\citep{suhavi-etal-2022-twitter-stmhd}.
However, these datasets are typically studied in isolation, making it challenging for patient simulation approaches to jointly capture clinically meaningful symptom semantics, counseling interaction dynamics, and long-term temporal evolution.

% =========================
\section{Conclusion}

In this work, we propose \textsc{Deprofile}, a data-grounded framework addressing key limitations in patient simulation. By aligning clinical dialogues with longitudinal social media via a two-stage process, we construct unified profiles with real-world temporal grounding. Our \textbf{Chain-of-Change (CoC) Agent} further transforms unstructured streams into verifiable structured memories. Both automatic metrics and expert evaluations demonstrate that \textsc{Deprofile} significantly improves realism and event richness over SOTA baselines. These results highlight the essential role of structured timelines and symptoms in suppressing hallucinations and ensuring clinical coherence.

\section*{Limitations}
Despite the significant progress made by Deprofile, this study has several limitations: First, our social media data is primarily sourced from Twitter-STMHD. Beyond demographic skew, we acknowledge that social media data requires ongoing calibration against established clinical cohorts to ensure the frequency and intensity of reported life events align with medical expectations. Future work will investigate weighting mechanisms to adjust for the positivity bias or outrage bias common in social platforms, further refining the authenticity of the simulated patient experiences.
Second, although we incorporated automated metrics such as G-Eval and MentalBERT, psychiatric diagnosis heavily relies on clinicians' intuition and non-verbal cues. Current text-only simulations cannot yet fully replicate these subtle interactive features.

\section*{Ethical Considerations}
Given that this study involves sensitive mental health data and patient simulation, we strictly adhere to the following ethical principles: First, all social media and counseling data used in this study underwent rigorous de-identification processes to remove personally identifiable information (PII). Second, \textsc{Deprofile} is intended solely as a simulated teaching tool to assist in psychiatrist training and counseling simulation. It is strictly prohibited for use in real clinical diagnosis. We urge the community to exercise high caution when deploying such technologies.

\bibliography{custom.bib}

@article{cohen2025evaluating,
  title={Evaluating the LLM-simulated Impacts of Big Five Personality Traits and AI Capabilities on Social Negotiations},
  author={Cohen, Myke C and Kao, Hsien-Te and Nguyen, Daniel and Lynch, Spencer and Volkova, Svitlana and Su, Zhe and Sap, Maarten},
  year={2025}
}

@article{Maike2022StressfulLifeEvents,
  title   = {Stressful Life Events Increase the Risk of Major Depressive Episodes: A Population-Based Twin Study},
  author  = {L{\o}v{\aa}s, Maike E. S. and R{\o}ysamb, Espen and Kendler, Kenneth S. and Czajkowski, Nils O. and Reichborn-Kjennerud, Ted and Ystr{\o}m, Eivind},
  journal = {Psychological Medicine},
  year    = {2022},
  volume  = {54},
  number  = {10},
  pages   = {1802--1812},
  doi     = {10.1017/S0033291722002227},
  issn    = {0033-2917},
  url     = {https://pmc.ncbi.nlm.nih.gov/articles/PMC10476058/}
}

@misc{zhai2024chinesementalbert,
  title        = {Chinese Mental{BERT}: Domain-Adaptive Pre-training on Social Media for Chinese Mental Health Text Analysis},
  author       = {Wei Zhai and Hongzhi Qi and Qing Zhao and Jianqiang Li and Ziqi Wang and Han Wang and Bing Xiang Yang and Guanghui Fu},
  year         = {2024},
  eprint       = {2402.09151},
  archivePrefix = {arXiv},
  primaryClass = {cs.CL},
  url          = {https://arxiv.org/abs/2402.09151}
}

@inproceedings{zhang2023geval,
  title     = {G-Eval: {NLG} Evaluation using {GPT-4} with Better Human Alignment},
  author    = {Zhang, Yuhui and Tang, Tianyi and Wang, Xiaoyang and others},
  booktitle = {Proceedings of the 2023 Conference on Empirical Methods in Natural Language Processing},
  year      = {2023},
  archivePrefix = {arXiv},
  eprint    = {2303.16634}
}

@article{Yarrington2023LifeEvents,
  title   = {The Role of Positive and Negative Aspects of Life Events in Depressive and Anxiety Symptoms},
  author  = {Yarrington, Julia S. and Metts, Allison V. and Zinbarg, Richard E. and Nusslock, Robin and Wolitzky-Taylor, Kate and Hammen, Constance L. and Kelley, Nicholas J. and Bookheimer, Susan and Craske, Michelle G.},
  journal = {Clinical Psychological Science},
  year    = {2023},
  volume  = {11},
  number  = {5},
  pages   = {910--920},
  doi     = {10.1177/21677026221141654},
  issn    = {2167-7034},
  url     = {https://pmc.ncbi.nlm.nih.gov/articles/PMC10530959/}
}

@article{dai2025profile,
  title={Profile-LLM: Dynamic Profile Optimization for Realistic Personality Expression in LLMs},
  author={Dai, Shi-Wei and Shie, Yan-Wei and Yang, Tsung-Huan and Ku, Lun-Wei and Li, Yung-Hui},
  journal={arXiv preprint arXiv:2511.19852},
  year={2025}
}

@inproceedings{wang-etal-2024-incharacter,
    title = "{I}n{C}haracter: Evaluating Personality Fidelity in Role-Playing Agents through Psychological Interviews",
    author = "Wang, Xintao  and
      Xiao, Yunze  and
      Huang, Jen-tse  and
      Yuan, Siyu  and
      Xu, Rui  and
      Guo, Haoran  and
      Tu, Quan  and
      Fei, Yaying  and
      Leng, Ziang  and
      Wang, Wei  and
      Chen, Jiangjie  and
      Li, Cheng  and
      Xiao, Yanghua",
    booktitle = "Proceedings of the 62nd Annual Meeting of the Association for Computational Linguistics (Volume 1: Long Papers)",
    month = aug,
    year = "2024",
    address = "Bangkok, Thailand",
    publisher = "Association for Computational Linguistics",
    url = "https://aclanthology.org/2024.acl-long.102/",
    doi = "10.18653/v1/2024.acl-long.102",
    pages = "1840--1873",
}

@inproceedings{suhavi-etal-2022-twitter-stmhd,
  author    = {Suhavi and
               Asmit Kumar Singh and
               Udit Arora and
               Somyadeep Shrivastava and
               Aryaveer Singh and
               Rajiv Ratn Shah and
               Ponnurangam Kumaraguru},
  title     = {Twitter-STMHD: An Extensive User-Level Database of Multiple Mental Health Disorders},
  booktitle = {Proceedings of the Sixteenth International {AAAI} Conference on Web and Social Media (ICWSM 2022)},
  year      = {2022},
  pages     = {1182--1191}
}

@inproceedings{yao-etal-2022-d4,
  title = "D4: a {C}hinese Dialogue Dataset for Depression-Diagnosis-Oriented Chat",
  author = "Yao, Binwei  and
    Shi, Chao  and
    Zou, Likai  and
    Dai, Lingfeng  and
    Wu, Mengyue  and
    Chen, Lu  and
    Wang, Zhen  and
    Yu, Kai",
  editor = "Goldberg, Yoav  and
    Kozareva, Zornitsa  and
    Zhang, Yue",
  booktitle = "Proceedings of the 2022 Conference on Empirical Methods in Natural Language Processing",
  month = dec,
  year = "2022",
  address = "Abu Dhabi, United Arab Emirates",
  publisher = "Association for Computational Linguistics",
  url = "https://aclanthology.org/2022.emnlp-main.156/",
  doi = "10.18653/v1/2022.emnlp-main.156",
  pages = "2438--2459"
}

@inproceedings{li-etal-2023-understanding,
  title = "Understanding Client Reactions in Online Mental Health Counseling",
  author = "Li, Anqi  and
    Ma, Lizhi  and
    Mei, Yaling  and
    He, Hongliang  and
    Zhang, Shuai  and
    Qiu, Huachuan  and
    Lan, Zhenzhong",
  editor = "Rogers, Anna  and
    Boyd-Graber, Jordan  and
    Okazaki, Naoaki",
  booktitle = "Proceedings of the 61st Annual Meeting of the Association for Computational Linguistics (Volume 1: Long Papers)",
  month = jul,
  year = "2023",
  address = "Toronto, Canada",
  publisher = "Association for Computational Linguistics",
  url = "https://aclanthology.org/2023.acl-long.577/",
  doi = "10.18653/v1/2023.acl-long.577",
  pages = "10358--10376"
}

@article{wang2024patient,
  title     = {{PATIENT-$\Psi$}: Using Large Language Models to Simulate Patients for Training Mental Health Professionals},
  author    = {Wang, Ruiyi and Milani, Federico L. and Wang, Zihao and Zhang, Ruijie and Wang, Yuhui and Zhao, Haoran and Zhao, Xuan and Zhu, Jiaxuan and Zhang, Xuan and Wang, Zihao},
  journal   = {arXiv preprint arXiv:2405.19660},
  year      = {2024}
}

@inproceedings{louie2024roleplay,
  title     = {Roleplay-doh: Enabling Domain-Experts to Create LLM Simulations through Human-LLM Collaboration},
  author    = {Louie, Ryan and Wang, Ruiyi and Milani, Federico L. and Wang, Zihao},
  booktitle = {Proceedings of the 2024 Conference on Empirical Methods in Natural Language Processing},
  year      = {2024},
  pages     = {10711--10729}
}

@inproceedings{wang2025annaagent,
  title     = {AnnaAgent: Dynamic Evolution Agent System with Multi-Session Memory for Psychological Counseling},
  author    = {Wang, Mingxuan and others},
  booktitle = {Findings of the Association for Computational Linguistics: ACL 2025},
  year      = {2025},
  pages     = {1192}
}

@article{liu2025eeyore,
  title     = {Eeyore: Realistic Depression Simulation via Supervised and Preference Optimization},
  author    = {Liu, Shuyang and others},
  journal   = {arXiv preprint arXiv:2503.00018},
  year      = {2025}
}

@misc{li2025patientsim,
  title     = {PatientSim: A Persona-Driven Simulator for Realistic Doctor-Patient Interactions},
  author    = {Li, Yuhang and Wang, Ruiyi and others},
  year      = {2025},
  url       = {https://arxiv.org/abs/2505.17818}
}

@misc{schick2024talkdep,
  title     = {TalkDep: Clinically Grounded LLM Personas for Conversation Simulation in Depression Diagnosis},
  author    = {Schick, Timo and others},
  year      = {2024},
  url       = {https://arxiv.org/abs/2508.04248}
}

@inproceedings{Chen2024MappingLC,
  title={Mapping Long-term Causalities in Psychiatric Symptomatology and Life Events from Social Media},
  author={Siyuan Chen and Meilin Wang and Minghao Lv and Zhiling Zhang and Juqianqian Juqianqian and Dejiyangla Dejiyangla and Yujia Peng and Ke Zhu and Mengyue Wu},
  booktitle={North American Chapter of the Association for Computational Linguistics},
  year={2024}
}

@inproceedings{Zhang2022SymptomIF,
  title={Symptom Identification for Interpretable Detection of Multiple Mental Disorders on Social Media},
  author={Zhiling Zhang and Siyuan Chen and Mengyue Wu and Ke Zhu},
  booktitle={Conference on Empirical Methods in Natural Language Processing},
  year={2022}
}

@inproceedings{jiang2024big5chat,
  title     = {BIG5-CHAT: Shaping {LLM} Personalities Through Training on Human Preference Data},
  author    = {Jiang, Albert Q. and Ren, Mark and others},
  booktitle = {Proceedings of the 2025 Conference of the North American Chapter of the Association for Computational Linguistics},
  year      = {2024}
}

@misc{azam2024llms,
  title     = {How Effective are Large Language Models (LLMs) at Inferring Big Five Personality Traits from Text?},
  author    = {Azam, Muhammad and others},
  year      = {2024},
  url       = {https://ceur-ws.org/Vol-3962/paper15.pdf}
}

\appendix

\section{ Prompts for Demographic Inference}
\label{promptappendixa}
\subsection{Gender Inference Prompt}
% \begin{displayquote}
You will receive a post that may allow you to infer the user's gender. Please determine whether the user's gender can be identified according to the following rules, and output a single Python string. Choose one from \texttt{["M", "F", "Unknown"]}, where \texttt{"M"} represents male, \texttt{"F"} represents female, and \texttt{"Unknown"} means indeterminable. Only make a judgment when the post clearly indicates that the user identifies as male or female; otherwise, output "Unknown". \\
\textbf{Examples:} \\
Example 1: User post: "I am not a good daughter...." $\rightarrow$ Judgment: \texttt{"F"}. The user identifies as a daughter, so they consider themselves female. \\
Example 2: User post: "I really like wearing long dresses....." $\rightarrow$ Judgment: \texttt{"Unknown"}. Liking dresses does not necessarily indicate the user identifies as female.
% \end{displayquote}

\subsection{Age Inference Prompt}
% \begin{displayquote}
You will receive a post that may provide clues about a user's age. Please determine whether the user's age can be inferred and output a JSON object according to the following rules. Choose one from \texttt{["0-17", "18-25", "26-35", "36-50", "50+", "Unknown"]}, which correspond to: minors, college-aged youth, young professionals, middle-aged adults, seniors, and cases where the content provides no information about age. \\
\textbf{Examples:} \\
Example 1: User post: "Today is my 22nd birthday..." $\rightarrow$ Judgment: \texttt{"18-25"}. \\
Example 2: User post: "Last year I retired from Amazon..." $\rightarrow$ Judgment: \texttt{"50+"}.
% \end{displayquote}

\subsection{Employment Status Inference Prompt}
% \begin{displayquote}
You will receive a post that may allow you to infer the user's work status. Please determine the user's employment status according to common sense and output a single Python string. Choose one from \texttt{["employed", "unemployed", "student", "retired", "Unknown"]}, where:
\begin{itemize}
    \item \texttt{"employed"}: means the user currently has a job.
    \item \texttt{"unemployed"}: means not currently working.
    \item \texttt{"student"}: means the user is studying in school or university.
    \item \texttt{"retired"}: means the user has stopped working due to age.
    \item \texttt{"Unknown"}: means indeterminable.
\end{itemize}
Make your judgment based on reasonable real-world understanding of occupational expressions in the post; if the status still cannot be inferred, output "Unknown". \\
\textbf{Examples:} \\
Example 1: User post: "Just got a promotion at work today!" $\rightarrow$ Judgment: \texttt{"employed"}. The user mentions a promotion, which implies they are currently employed. \\
Example 2: User post: "Graduation ceremony next week!" $\rightarrow$ Judgment: \texttt{"student"}. The user mentions graduation, indicating they are (or were recently) a student. \\
Example 3: User post: "I lost my job last month and still can't find one...!" $\rightarrow$ Judgment: \texttt{"unemployed"}. The user states they lost their job and are not working, indicating unemployed status. \\
Example 4: User post: "Enjoying my retirement life by the beach." $\rightarrow$ Judgment: \texttt{"retired"}. The user mentions retirement, clearly indicating retired status.
% \end{displayquote}

\subsection{Marital Status Inference Prompt}
% \begin{displayquote}
You will receive a post that may allow you to infer the user's marital status. Please determine the user's marital status according to common sense and output a single Python dictionary. Choose one from \texttt{["married", "single", "divorced", "widowed", "Unknown"]}, where:
\begin{itemize}
    \item \texttt{"married"}: means the user is married.
    \item \texttt{"single"}: means unmarried.
    \item \texttt{"divorced"}: means separated from a previous spouse.
    \item \texttt{"widowed"}: means their spouse has passed away.
    \item \texttt{"Unknown"}: means indeterminable.
\end{itemize}
Make your judgment based on reasonable real-world understanding of marital expressions in the post; if the status still cannot be inferred, output "Unknown". \\
\textbf{Examples:} \\
Example 1: User post: "I just celebrated my 5th wedding anniversary today!" $\rightarrow$ Judgment: \texttt{"married"}. The user mentions a wedding anniversary, indicating they are married. \\
Example 2: User post: "It's been two years since my husband passed away..." $\rightarrow$ Judgment: \texttt{"widowed"}. The user mentions their husband's death, indicating widowed status. \\
Example 3: User post: "My girlfriend broke up with me last week..." $\rightarrow$ Judgment: \texttt{"single"}. The user says they broke up, indicating they are now single. \\
Example 4: User post: "I finalized my divorce last month." $\rightarrow$ Judgment: \texttt{"divorced"}. The user mentions finalizing a divorce, indicating divorced status.
% \end{displayquote}

\section{Chain-of-Change (CoC) Timeline Agent}
\label{app:coc-detail}

\subsection{Motivation}
Patient simulation over multiple turns requires consistent references to past symptoms and events, especially when a doctor asks time-specific follow-up questions (e.g., \emph{``When did it start?''}, \emph{``What happened before that?''}). Naively providing raw timeline text often leads to vague temporal expressions or hallucinated details. We therefore introduce a Chain-of-Change (CoC) Timeline Agent that converts raw day-indexed timelines into a \textbf{verifiable} and \textbf{query-friendly} memory representation, enabling (i) precise time-point referencing and (ii) structured change tracking across episodes.

\subsection{Inputs and Notation}
Each profile contains two timeline streams:
\begin{itemize}
    \item \textbf{Symptom timeline} $\mathcal{T}_{s}=\{(t_i, y_i, x_i)\}$ where $t_i \in \mathbb{Z}$ is a day index, $y_i$ is a symptom label, and $x_i$ is a post/tweet text.
    \item \textbf{Life-event timeline} $\mathcal{T}_{e}=\{(t_j, z_j, x_j)\}$ where $t_j$ is a day index, $z_j$ is an event label (if available), and $x_j$ is a post/tweet text.
\end{itemize}
For a given dialogue state, we define an \textbf{anchor day} $t^{\star}$ (default: the latest timestamp in the timeline) and a lookback horizon $H$ (e.g., $H=90$ days). We keep items whose timestamps satisfy $t \in (t^{\star}-H, t^{\star}]$.

\subsection{CoC Memory: Dual-Channel Representation}
CoC outputs a dual-channel memory:
\begin{enumerate}
    \item \textbf{Graph memory} $\mathcal{G}=(\mathcal{V},\mathcal{E})$ for alignment and evaluation.
    \item \textbf{Rendered cards} $\mathcal{C}$ (episode summaries) for downstream generation.
\end{enumerate}

\paragraph{Time normalization.}
For each item with timestamp $t$, we compute:
\[
\texttt{days\_ago} = t^{\star}-t,\quad
\texttt{relative\_cn} = f(\texttt{days\_ago})
\]
where $f(\cdot)$ maps day offsets to a coarse natural-language relative time string (e.g., ``\(\texttt{2 weeks ago}\)'').

\paragraph{Nodes.}
For each symptom item $(t_i,y_i,x_i)$, we create a symptom node:
\[
v_i = \{\texttt{id},\texttt{type}=\texttt{Symptom},\texttt{symptom}=y_i,
\]
\[
\texttt{evidence}=x_i,\texttt{time\_norm}\}
\]
Optionally, we call an instruction model to extract a \textbf{symptom triple} and a short summary from $x_i$ while constraining it to be consistent with $y_i$:
\[
\texttt{symptom\_triple} \leftarrow \textsc{LLMExtractSymptom}(y_i, x_i)
\]
For each event item $(t_j,z_j,x_j)$, we create a life-event node:

and optionally extract an \textbf{event triple} and summary:
\[
\texttt{event\_triple} \leftarrow \textsc{LLMExtractEvent}(z_j, x_j)
\]
In both cases, the extractor must return \texttt{None} if the post does not describe a meaningful first-person experience, preventing noisy/hallucinated nodes.

\paragraph{Edges.}
We construct a basic temporal skeleton by sorting nodes by \texttt{timestamp\_day} and adding:
\begin{itemize}
    \item \textbf{Temporal edges:} For any two nodes $v, w$ where $v$ temporally precedes $w$, we add an edge:
    \[
        (v \rightarrow w) \in \mathcal{E} 
    \]
    \[
    \quad \text{with relation } \texttt{temporal\_precedes}
    \]
    \item \textbf{Persistence edges:} For symptom nodes with the same label $y$ at consecutive time steps $k$ and $k+1$, we add an edge:
    \[
        (v_{y,k} \rightarrow v_{y,k+1}) \in \mathcal{E} \quad \text{with relation } \texttt{persists}
    \]
\end{itemize}
(Extensions like \texttt{triggers}/\texttt{worsens}/\texttt{alleviates} can be added when causal evidence is available.)

\subsection{Episode Aggregation and Rendered Cards}
To improve usability by downstream LLMs, we aggregate nodes into fixed-size windows of $W$ days (e.g., $W=7$). Each node is assigned to an episode bucket by:
\[
\texttt{bucket}(v)=\left\lfloor \frac{\texttt{days\_ago}(v)}{W} \right\rfloor
\]
For each bucket, we render a short card containing (i) the episode time range and (ii) salient symptom/event content. Crucially, each card includes a \textbf{representative time point} in the format:
\[
\texttt{represent\_time}: 
\]
\[
\ \texttt{relative\_cn}(\texttt{days\_ago}) \ \text{(}\texttt{days\_ago})
\]
Rendered cards are produced by deterministic templates (not an LLM) to avoid introducing new facts and to ensure reproducibility.

\subsection{Downstream Usage Protocol (Temporal Faithfulness Constraint)}
When the patient simulator references timeline content, we enforce:
\begin{itemize}
    \item \textbf{Card-first referencing}: prefer citing rendered cards over raw graph JSON.
    \item \textbf{Mandatory time-point}: each referenced timeline statement must include the card's \texttt{represent\_time} (e.g., ``14 days ago'').
    \item \textbf{No fabrication}: do not introduce dates/events/details not present in the cards/graph.
\end{itemize}
This protocol supports doctor follow-up queries about prior events by enabling precise temporal grounding.

\subsection{Algorithm Sketch}
\begin{algorithm}[h]
\caption{CoC Timeline Agent (Graph + Cards)}
\label{alg:coc}
\begin{algorithmic}[1]
\Require timeline stream $\mathcal{T}$, anchor day $t^{\star}$, horizon $H$, window size $W$
\State $\mathcal{T}' \leftarrow \{(t,\cdot,\cdot)\in\mathcal{T}: t\in(t^{\star}-H, t^{\star}]\}$
\State Initialize graph $\mathcal{G}=(\mathcal{V},\mathcal{E})$
\For{each item in $\mathcal{T}'$}
    \State Compute \texttt{time\_norm} including \texttt{days\_ago} and \texttt{relative\_cn}
    \State Create node; optionally run LLM extractor; skip if extractor returns \texttt{None}
    \State Add node to $\mathcal{V}$ and index by day/label
\EndFor
\State Add \texttt{temporal\_precedes} edges by sorting nodes by timestamp
\If{symptom stream}
    \State Add \texttt{persists} edges for repeated symptom labels
\EndIf
\State Bucket nodes into episodes by $\lfloor \texttt{days\_ago}/W \rfloor$
\State Render cards $\mathcal{C}$ for each episode using deterministic templates
\State \Return $\{\mathcal{G}, \mathcal{C}\}$
\end{algorithmic}
\end{algorithm}

\subsection{Implementation Notes}
We store rendered cards on disk as JSON:
\texttt{cards} is a list of \{\texttt{episode\_id}, \texttt{time\_range}, \texttt{card\_cn}\}.
Downstream prompt construction loads the corresponding file by \texttt{profile\_id} and \texttt{tweet\_user\_id} and appends (i) a temporal faithfulness rule and (ii) the card list to the system context.

\section{Prompt Components}
\label{app:prompt-components}
\noindent\textbf{B: Basic Demographic Profile.}\;
{\small
\begin{quote}\ttfamily
Use ONLY the following demographic profile:\newline
Age: <AGE>\newline
Gender: <GENDER>\newline
Employment status: <WORK\_STATUS>\newline
Marital status: <MARITAL\_STATUS>
\end{quote}
}

\noindent\textbf{R: Mental Disorder Related Risk (translated from implementation).}\;
{\small
\begin{quote}\ttfamily
You currently have <DEPRESSION\_RISK\_LEVEL> risk of depression.\newline
You currently have <SUICIDE\_RISK\_LEVEL> risk of suicide.\newline
\newline
Although you may have some symptoms and negative life events, you must stay
consistent with the specified depression severity and suicide risk. If you are
asked to fill in any scale, answer according to the condition described here.
\end{quote}
}

\noindent\textbf{P: Personality Traits (Big Five Traits).}\;
{\small
\begin{quote}\ttfamily
Follow the Big Five personality instructions below. Make decisions and speak
in this style throughout the conversation, but do NOT explicitly mention
psychological trait terms.\newline
\newline
Openness:\newline
- High: Be imaginative; use metaphors; embrace novelty and abstraction.\newline
- Medium: Balance pragmatism and innovation.\newline
- Low: Be traditional; focus on established facts; avoid abstract/vague ideas.\newline
\newline
Conscientiousness:\newline
- High: Be highly disciplined and detail-oriented.\newline
- Medium: Be reliable and organized, with flexibility when appropriate.\newline
- Low: Be casual and unstructured; may appear procrastinating.\newline
\newline
Extraversion:\newline
- High: Be enthusiastic and energetic; actively lead the conversation.\newline
- Medium: Communicate smoothly without dominating.\newline
- Low: Be reserved; answer briefly; avoid initiating topics.\newline
\newline
Agreeableness:\newline
- High: Be gentle and accommodating; avoid conflicts.\newline
- Medium: Be friendly but with boundaries.\newline
- Low: Be blunt; focus on facts; be skeptical/critical.\newline
\newline
Neuroticism:\newline
- High: Show anxiety and mood swings; worry easily.\newline
- Medium: Show normal emotional reactivity and recovery.\newline
- Low: Be unshakably calm and emotionally stable.
\end{quote}
}

\noindent\textbf{S: Symptom Attributes (prompt structure; translated).}\;
{\small
\begin{quote}\ttfamily
Below are the patient's positive symptom attributes. Some positive attributes
are associated with timeline time points; when asked about these attributes,
mention the time points explicitly.\newline
\newline
<POSITIVE\_ATTRIBUTE\_DESCRIPTIONS\allowbreak\_WITH\_OPTIONAL\_TIMELINE>\newline
\newline
Other symptom timeline information (if available):\newline
<EXTRA\_TIMELINE\_SNIPPETS>\newline
\newline
Below are descriptions of the patient's normal conditions. If you are asked
about any of these symptoms/attributes, you must deny them.\newline
\newline
<NEGATIVE\_ATTRIBUTE\_DESCRIPTIONS>
\end{quote}
}

\noindent\textbf{A: Assessment Dialogue Data.}\;
{\small
\begin{quote}\ttfamily
Assessment-style reference (style only): Use the following diagnostic interview
snippets only to imitate answering style (brief answers, vague frequency/severity,
more details only when probed).\newline
\newline
<ASSESSMENT\_DIALOGUE\_SNIPPETS>
\end{quote}
}

\noindent\textbf{C: Counseling Dialogue Data.}\;
{\small
\begin{quote}\ttfamily
Counseling-style reference (style only): Use the following counseling snippets
only to imitate interaction style (tone, hesitation, avoidance, emotional expression).\newline
\newline
<COUNSELING\_DIALOGUE\_SNIPPETS>
\end{quote}
}

\noindent\textbf{T: Life Event Timeline (cards).}\;
{\small
\begin{quote}\ttfamily
Use the Life-Event Timeline Cards below to describe changes in mood and functioning.\newline
Rules:\newline
1) Only use events/impacts that appear in the cards; do not invent details.\newline
2) Include the time expression from the card when mentioning an event/impact.\newline
3) Only claim causality if both the event and the change are supported by the cards.\newline
4) Cite at most 1--2 cards per answer; avoid list-like narration.\newline
\newline
<LIFE\_EVENT\_CARDS>
\end{quote}
}

\section{G-Eval Prompt Format}
\label{app:g-eval}

The following prompt was designed for the GPT-5 evaluator to assess the quality of the simulated patient responses.

% 使用 quote 环境让整个 prompt 块向内缩进，区别于正文
\begin{quote}
\small % 可以稍微把字体调小一点，让内容更紧凑

You are a clinical psychology expert and an expert evaluator for role-playing dialogue systems. Your task is to evaluate the quality of the patient's responses in the dialogue based on (i) the patient profile and constraints, and (ii) the exact System Prompt that the patient model actually saw when generating its responses.

\vspace{1em} % 增加一点垂直间距
\textbf{[1. System Prompt seen by the patient model during generation (IMPORTANT): includes patient profile and constraints]}

% 使用 \texttt{} 来表示占位符，显示为打字机字体
\texttt{\{patient\_prompt\_text\}}

\vspace{1em}
\textbf{[2. Dialogue context (multi-turn; question = clinician/doctor, answer = patient)]}

\texttt{\{dialogue\}}

\vspace{1em}
\textbf{[3. Rating dimensions (each must be an integer from 1 to 5)]}

% 使用 description 环境来列出定义项
\begin{description}
    \setlength\itemsep{0.5em} % 设置列表项之间的间距
    \item[A) persona\_faithfulness:] Whether the responses are consistent with the demographics and Big Five personality traits (no contradictions; consistent speaking style).
    \item[B) event richness and temporal diversity:] Whether the dialogue mentions life events more concretely, with richer categories and more diverse time spans. Mentioning explicit time points should receive higher scores.
    \item[C) symptom\_consistency:] Whether the symptoms expressed in the dialogue are consistent with the profile's positive/negative symptoms, without contradictions or adding major symptoms not supported by the profile.
\end{description}

\vspace{1em}
\textbf{[4. Output requirements]}

% 使用 itemize 环境做无序列表
\begin{itemize}
    \setlength\itemsep{0em} % 让列表更紧凑
    \item Output STRICT JSON only. No Markdown. No extra text.
    \item Provide a score for each dimension, plus brief reasons and evidence points.
    \item Also extract a list of ``explicitly mentioned life events'' in the dialogue as \texttt{extracted\_events} (output an empty array if none).
\end{itemize}

The required JSON schema is:

\texttt{\{SCHEMA\_JSON\}}

\end{quote}

\section{Human Evaluation Details}
\label{app:human-eval}

\subsection{Expert Recruitment and Compensation}
To ensure the clinical validity and professional rigor of our evaluation, we recruited eight psychiatric experts to serve as human raters. The recruitment and compensation protocols were strictly governed by the following criteria:

\begin{itemize}
    \item \textbf{Expert Qualifications}: Each recruited expert has completed at least five years of professional higher education in clinical psychology or psychiatry. This specialized background ensures that the evaluators possess the domain expertise required to assess complex dimensions such as \textit{Symptom Consistency}, \textit{Clinical Realism}, and \textit{Psychological Plausibility}.
    \item \textbf{Ethical Compensation}: In accordance with ethical labor practices and to recognize the high level of specialized expertise provided, all experts were remunerated at a rate exceeding 200\% of the local statutory minimum wage. 
    \item \textbf{Anonymity and Informed Consent}: All participation was voluntary and based on informed consent. To maintain the integrity of the double-blind review process, recruitment and payment were handled through channels independent of the core research team, and all evaluator data were de-identified before analysis.
\end{itemize}

\subsection{Evaluation Guidelines}
The following guidelines were provided to expert reviewers for the human evaluation. Each reviewer receives a patient profile (\texttt{profile.txt}) and five dialogue outputs (from basic, basic+S, \textsc{Patient-$\Psi$}, \textsc{Eeyore}, and \textsc{Deprofile}). They rate each dialogue on a 1--5 scale (1 lowest, 5 highest) along five dimensions:

\begin{description}
    \setlength\itemsep{0.5em}
    \item[Realism] Does the response sound like a real person speaking? Are there concrete details, colloquial expressions, and natural emotional variation, or does it feel template-like?
    \item[Persona Faithfulness] Are age, gender, occupation, marital status, and personality traits consistent with the profile? Any contradictions?
    \item[Event Richness] Are specific life events mentioned? Are event types diverse (academic, interpersonal, family, work)? Is there a clear temporal structure (e.g., ``three months ago'', ``last week'')?
    \item[Symptom Consistency] Do the symptoms expressed in the dialogue align with the profile description? Overall impression suffices.
    \item[Overall] A holistic score synthesizing the above dimensions.
\end{description}

Reviewers are instructed to differentiate between models when possible and to avoid giving identical scores across all systems for the same sample.

\section{Symptom Attribute List}
\label{app:appendix_symptoms}
\subsection{Assessment Dialogue Symptom Attributes}

\begin{itemize}[noitemsep,topsep=0pt]
  \item sleep-light sleep
  \item sleep-difficulty falling asleep
  \item sleep-frequent dreaming
  \item sleep-early awakening
  \item sleep-sleep disturbance
  \item sleep-reduced sleep duration

  \item appetite-binge eating
  \item appetite-significant weight change
  \item appetite-loss of appetite
  \item appetite-appetite disturbance

  \item suicide-hopelessness
  \item suicide-suicidal ideation
  \item suicide-low self-worth
  \item suicide-self-blame
  \item suicide-self-harm tendency
  \item suicide-suicide attempt

  \item screening-mania
  \item screening-family history

  \item somatic-somatic discomfort
  \item somatic-psychomotor agitation
  \item somatic-psychomotor retardation

  \item emotion-depressed mood
  \item emotion-depressed mood over two weeks
  \item emotion-diurnal variation

  \item interest-loss of interest
  \item interest-scope all activities
  \item interest-emotional blunting
  \item interest-cause
  \item interest-loss of interest over two weeks
  \item interest-scope past hobbies

  \item social functioning-difficulty in daily life
  \item social functioning-difficulty in study or work
  \item social functioning-avoid social contact
  \item social functioning-avoid support from family or friends

  \item mental state-fatigue
  \item mental state-memory decline
  \item mental state-lack of confidence
  \item mental state-indecisiveness
  \item mental state-inattention
\end{itemize}
\subsection{Paired Symptom Attributes}

\begin{itemize}[noitemsep,topsep=0pt]
  \item \textit{Decreased Energy Tiredness Fatigue} $\leftrightarrow$ mental state-fatigue
  \item \textit{Depressed Mood} $\leftrightarrow$ emotion-depressed mood
  \item \textit{Inattention} $\leftrightarrow$ mental state-inattention
  \item \textit{Indecisiveness} $\leftrightarrow$ mental state-indecisiveness
  \item \textit{Suicidal Ideas} $\leftrightarrow$ suicide-suicidal ideation
  \item \textit{Worthlessness And Guilty} $\leftrightarrow$ suicide-low self-worth
  \item \textit{Diminished Emotional Expression} $\leftrightarrow$ interest-emotional blunting
  \item \textit{Drastical Shift In Mood And Energy} $\leftrightarrow$ screening-mania
  \item \textit{Loss Of Interest Or Motivation} $\leftrightarrow$ interest-loss of interest
  \item \textit{Pessimism} $\leftrightarrow$ suicide-hopelessness
  \item \textit{Poor Memory} $\leftrightarrow$ mental state-memory decline
  \item \textit{Sleep Disturbance} $\leftrightarrow$ sleep-sleep disturbance
  \item \textit{Hyperactivity Agitation} $\leftrightarrow$ somatic-psychomotor agitation
  \item \textit{Catatonic Behavior} $\leftrightarrow$ somatic-psychomotor retardation
  \item \textit{Fear About Social Situations} $\leftrightarrow$ social functioning-avoid social contact
  \item \textit{Weight And Appetite Change} $\leftrightarrow$ appetite-appetite disturbance
  \item \textit{Fear Of Gaining Weight} $\leftrightarrow$ appetite-significant weight change
\end{itemize}
\subsection{Social Media--Only Symptom Attributes}
The following attributes appear only in social media data.
\begin{itemize}[noitemsep,topsep=0pt]
  \item \textit{Anxious Mood}
  \item \textit{Autonomic Symptoms}
  \item \textit{Cardiovascular Symptoms}
  \item \textit{Gastrointestinal Symptoms}
  \item \textit{Genitourinary Symptoms}
  \item \textit{Respiratory Symptoms}
  \item \textit{Impulsivity}
  \item \textit{Avoidance Of Stimuli}
  \item \textit{Compensatory Behaviors To Prevent Weight Gain}
  \item \textit{Compulsions}
  \item \textit{Fears Of Being Negatively Evaluated}
  \item \textit{Flight Of Ideas}
  \item \textit{Intrusion Symptoms}
  \item \textit{More Talktive}
  \item \textit{Obsession}
  \item \textit{Obsessions}
  \item \textit{Panic Fear}
  \item \textit{Do Things Easily Get Painful Consequences}
  \item \textit{Anger Irritability}
\end{itemize}

\section{Example Deprofile Instance}
\label{app:deprofile_example}
Table~\ref{tab:deprofile_example} shows an example patient profile constructed by
the \textsc{Deprofile} framework, illustrating how multi-source information is
unified into a structured representation.

\begin{table}[hbp]
\caption{An example patient profile constructed by the \textsc{Deprofile} framework.}
\label{tab:deprofile_example}
\centering
\small
\setlength{\tabcolsep}{6pt}
\begin{tabular}{p{0.34\columnwidth} p{0.62\columnwidth}}
\toprule
\textbf{Attribute} & \textbf{Value} \\
\midrule
Demographics &
24-year-old male; single; employed \\

Personality&
Openness: 5;\;
Conscientiousness: 4;\;
Extraversion: 3;\;
Agreeableness: 6;\;
Neuroticism: 6 \\

Longitudinal Alignment &
Matched social media timeline (similarity = 0.96;
symptom similarity = 0.67) \\

Positive Symptoms &
Suicidal ideation; hopelessness; low self-worth; fatigue;
sleep disturbance; appetite dysregulation; psychomotor agitation/retardation;
social withdrawal; indecisiveness \\

Negative Symptoms &
No suicide attempt; no severe work impairment;
no persistent sleep disorder \\

Clinical Summary &
Persistent depressed mood, anhedonia, low energy, and cognitive difficulties
over the past month, indicating a moderate-to-severe depressive condition \\

Risk Assessment &
Depression risk: 3;\; Suicide risk: 0 \\
\bottomrule
\end{tabular}

\end{table}

\onecolumn

\section{Full Three-stage Question Lists.}
\label{appendixB}
For reproducibility, the complete question lists for all three stages (exact wording and their attribute/category mapping) are provided here.

Table \ref{tab:question_list} presents the complete list of questions used in our dataset construction and evaluation. The questions are categorized by the specific module or psychological dimension they target.

% Make sure to include \usepackage{longtable} and \usepackage{booktabs} in your preamble

\begin{small}
\begin{longtable}{p{0.15\textwidth} p{0.3\textwidth} p{0.5\textwidth}}
\caption{The structured question list used for dialogue generation and assessment.} \label{tab:question_list} \\
\toprule
\textbf{Category} & \textbf{Topic / Label} & \textbf{Question (Translated)} \\
\midrule
\endfirsthead

\multicolumn{3}{c}%
{{\bfseries \tablename\ \thetable{} -- continued from previous page}} \\
\toprule
\textbf{Category} & \textbf{Topic / Label} & \textbf{Question (Translated)} \\
\midrule
\endhead

\midrule
\multicolumn{3}{r}{{Continued on next page}} \\
\bottomrule
\endfoot

\bottomrule
\endlastfoot

% --- Profile ---
\textbf{Profile} 
 & Intro & How old are you? \\
 & Bio & What is your gender? \\
 & Profession & Are you currently working or studying? What is your major or field? \\
 & Marital Status & What is your current relationship status? Are you married or in a relationship? \\
\midrule

% --- Big Five ---
\textbf{Big Five} 
 & Extraversion (Talkativeness) & Tell me, are you generally a talkative person, or do you prefer to stay quiet? \\
 & Agreeableness (Fault Finding) & When interacting with others, do you tend to find fault or pick at their mistakes? \\
 & Conscientiousness (Thoroughness) & When you do things, do you strive to be thorough and leave no stone unturned? \\
 & Neuroticism (Depression) & Have you felt depressed, down, or melancholic recently? \\
 & Openness (New Ideas) & Are you someone who likes to come up with novel ideas through imagination? \\
\midrule

% --- Sleep ---
\textbf{Sleep} 
 & Light Sleep & Do you wake up easily after falling asleep, or do you sleep soundly? \\
 & Difficulty Falling Asleep & After lying in bed, how long does it usually take to actually fall asleep? Do you toss and turn for a long time? \\
 & Dreaming & Have you been dreaming a lot recently? Do you feel tired upon waking, as if you hadn't slept? \\
 & Early Awakening & Do you wake up earlier than your alarm? Can you go back to sleep afterwards? \\
 & General Quality & Overall, how would you rate your sleep quality recently? Is there anything bothering you? \\
 & Duration & On average, how many hours do you actually sleep per night now? \\
\midrule

% --- Appetite ---
\textbf{Appetite} 
 & Binge Eating & When you are in a bad mood recently, do you uncontrollably eat a lot? \\
 & Weight Change & Has there been any significant weight gain or loss recently? Roughly by how much? \\
 & Loss of Appetite & Do you still have an appetite for foods you usually like, or do you feel like eating nothing? \\
 & General Status & How has your eating been? Any changes compared to before? \\
\midrule

% --- Suicide Risk ---
\textbf{Suicide Risk} 
 & Hopelessness & Given your current situation, do you have confidence in the future, or do you feel hopeless? \\
 & Ideation & When the pain is intense, have you had thoughts of ending your life? \\
 & Low Self-worth & Do you often feel useless or like a burden to others? \\
 & Guilt & Do you often blame yourself for things and feel it's all your fault? \\
 & Self-harm & When in extreme distress, have you done anything to hurt your body to vent? \\
 & Behavior/Attempts & Beyond thoughts, have you ever actually carried out a suicide plan or made an attempt? \\
\midrule

% --- Screening ---
\textbf{Screening} 
 & Mania & Have you ever had a period where you felt extremely energetic, your mind raced, and you needed no sleep? \\
 & Family History & Have any family members, like parents or siblings, had similar emotional issues or seen a psychiatrist? \\
\midrule

% --- Somatic ---
\textbf{Somatic} 
 & Physical Discomfort & Besides mood, do you have physical discomfort? Like headaches, stomachaches, or palpitations? \\
 & Psychomotor Agitation & Do you feel flustered or restless, feeling the need to walk around to feel better? \\
 & Psychomotor Retardation & Do you feel slowed down recently? Like speaking, walking, or thinking is slower than before? \\
\midrule

% --- Mood ---
\textbf{Mood} 
 & Depressed Mood & How is your mood recently? Do you feel unable to be happy? \\
 & Duration ($>$2 weeks) & How long has this low mood lasted? Has it been more than two weeks? \\
 & Diurnal Variation & Do you feel worse in the morning upon waking, or does it get harder in the evening? \\
\midrule

% --- Interest ---
\textbf{Interest} 
 & Loss of Interest & Can you still muster the energy to do hobbies you found interesting before? \\
 & Scope (All) & Is it a loss of interest in everything, or just some things? \\
 & Apathy & Do you feel anything regarding events around you or family care? Or do you feel numb? \\
 & Reason & What do you think caused you to lose interest in these things? \\
 & Duration ($>$2 weeks) & When did this state of disinterest start? Has it lasted over half a month? \\
 & Past Hobbies & Can you tell me specifically what you used to like doing? How does it feel thinking about them now? \\
\midrule

% --- Social/Mental ---
\textbf{Function \& Mental} 
 & Daily Difficulties & Do daily chores like bathing, eating, or cleaning feel difficult recently? \\
 & Work/Study Issues & Has this state affected your work efficiency? For example, can you finish tasks on time? \\
 & Fatigue & Do you feel physically exhausted, even if you haven't done much? \\
 & Social Avoidance & Do you feel like avoiding people recently? Including gatherings with friends or colleagues? \\
 & Memory Loss & Do you feel your memory has worsened? Like forgetting what you just said or where you put things? \\
 & Lack of Confidence & Do you feel confident when making decisions or handling matters? \\
 & Avoid Support & When unhappy, do you tell family or friends? Or do you tend to hide it from them? \\
 & Indecisiveness & Do you struggle to decide on simple things, like what to eat for lunch? \\
 & Concentration & Can you focus when reading or watching TV? Or does your mind wander easily? \\
\midrule

% --- Chit-chat ---
\textbf{Interaction} 
 & Seek Opinion & Regarding your current situation, would you like to know my opinion or suggestions? \\
 & Passive Info & If I didn't ask, would you prefer not to bring these things up? \\
 & Active Info & Besides what I asked, is there any other information about your state you feel you need to tell me? \\
 & Self-disclosure (Self) & Tell me about yourself, how do you evaluate yourself recently? \\
 & Self-disclosure (World) & Do you have any emotional reaction to recent news or events around you? \\
\midrule

% --- Timeline ---
\textbf{Timeline} 
 & Phase 1: Event & What special events happened in your life three months ago? \\
 & Phase 1: Reaction & What was your immediate reaction when this happened? \\
 & Phase 1: Impact & Did this event immediately impact your daily life back then? \\
 & Phase 2: Event & What special events happened in your life during the month in between? \\
 & Phase 2: Affect & How did this event affect your emotions and state? \\
 & Phase 2: Details & Can you detail the impact of this event on you? \\
 & Phase 3: Current & Back to the last week, how is your overall state? Did anything new happen? \\
 & Phase 3: Impact Check & In what ways have you been affected by this? \\
 & Phase 3: Capability & Do you find daily chores difficult these days? Can you detail the impact? \\
 & Phase 3: Future & How do you think your future life will be affected by previous events? \\

\end{longtable}
\end{small}

\newpage

\section{Full Results Tables}
\label{app:fullresults}
% in preamble (if not already):
% \usepackage{booktabs}
% \usepackage{multirow}
% \usepackage{array}
\begin{table*}[hbp]
\centering
\small
\setlength{\tabcolsep}{4pt}
\caption{Ablation study under the \textit{DeepSeek-V3.2-Exp} backbone.
We report 95\% confidence intervals (CI) for semantic realism and inter-patient diversity,
and mean $\pm$ standard deviation for G-Eval dimensions.
$\uparrow$ indicates higher is better.}
\label{tab:ablation_ci_geval}
\begin{tabular}{l|cc|ccc}
\toprule
\textbf{Model Variant}
& \textbf{Realism (95\% CI)} 
& \textbf{Diversity (95\% CI)} 
& \textbf{Event} 
& \textbf{Persona} 
& \textbf{Symptom} \\
\midrule

basic
& [0.9252, 0.9369]
& [0.0495, 0.0585]
& 4.60 $\pm$ 0.51
& 4.71 $\pm$ 0.64
& 3.99 $\pm$ 0.94 \\

basic+T
& [0.9396, 0.9484]
& [0.0515, 0.0661]
& 4.84 $\pm$ 0.40
& 4.63 $\pm$ 0.65
& 3.89 $\pm$ 0.87 \\

\midrule
basic+S
& [0.9304, 0.9414]
& [0.0556, 0.0661]
& 4.80 $\pm$ 0.43
& 4.94 $\pm$ 0.31
& 4.84 $\pm$ 0.42 \\

basic+ST
& [0.9363, 0.9448]
& [0.0577, 0.0781]
& \textbf{4.90} $\pm$ 0.32
& 4.80 $\pm$ 0.51
& 4.71 $\pm$ 0.55 \\

\midrule
full w/o T 
& \textbf{[0.9364, 0.9461]}
& \textbf{[0.0597, 0.0715]}
& 4.57 $\pm$ 0.55
& \textbf{4.97} $\pm$ 0.23
& \textbf{4.86} $\pm$ 0.38 \\

\textsc{Deprofile}
& [0.9337, 0.9445]
& [0.0570, 0.0687]
& 4.86 $\pm$ 0.37
& 4.87 $\pm$ 0.43
& 4.79 $\pm$ 0.45 \\

\bottomrule
\end{tabular}
\end{table*}

\begin{table*}[hbp]
\centering
\small
\setlength{\tabcolsep}{5pt}
\caption{Transposed view of Table~\ref{tab:main_results}. 
Rows correspond to evaluation metrics and methods, while columns correspond to backbone models.
All values are identical to those reported in Table~\ref{tab:main_results}.}
\label{tab:appendix_stats_transposed}
\begin{tabular}{l|cccc}
\toprule
\textbf{Metric / Method} 
& \textit{Llama-3.1-8B} 
& \textit{Qwen-3-4B} 
& \textit{GPT-4o-mini} 
& \textit{DeepSeek-V3.2} \\
\midrule

\multicolumn{5}{l}{\textbf{Realism (95\% CI)}} \\
\midrule
Patient-$\Psi$ 
& [0.8777, 0.8977] 
& [0.8828, 0.9016] 
& [0.9206, 0.9328] 
& [0.9147, 0.9310] \\

Eeyore 
& [0.8869, 0.9040] 
& [0.8901, 0.9053] 
& [0.9175, 0.9317] 
& [0.9148, 0.9285] \\

\textbf{Deprofile} 
& \textbf{[0.9308, 0.9415]} 
& \textbf{[0.9057, 0.9189]} 
& \textbf{[0.9345, 0.9444]} 
& \textbf{[0.9337, 0.9445]} \\

\midrule
\multicolumn{5}{l}{\textbf{Diversity (95\% CI)}} \\
\midrule
\textsc{Patient-$\Psi$ }
& [0.0792, 0.0903] 
& [0.0227, 0.0247] 
& [0.0207, 0.0321] 
& [0.0424, 0.0513] \\

\textsc{Eeyore }
& [0.0792, 0.0903] 
& [0.0189, 0.0222] 
& [0.0239, 0.0335] 
& [0.0409, 0.0473] \\

\textsc{Deprofile (Ours)} 
& \textbf{[0.0792, 0.0903]} 
& \textbf{[0.0365, 0.0419]} 
& \textbf{[0.0436, 0.0563]} 
& \textbf{[0.0570, 0.0687]} \\

\midrule
\multicolumn{5}{l}{\textbf{Event Richness (Mean $\pm$ Std)}} \\
\midrule
\textsc{Patient-$\Psi$ }
& 3.74 $\pm$ 0.47 
& 4.76 $\pm$ 0.48 
& 2.95 $\pm$ 0.50 
& 4.76 $\pm$ 0.48 \\

\textsc{Eeyore }
& 3.99 $\pm$ 0.45 
& 4.83 $\pm$ 0.42 
& 3.08 $\pm$ 0.57 
& 3.77 $\pm$ 0.66 \\

\textbf{Deprofile} 
& \textbf{4.24 $\pm$ 0.59} 
& \textbf{4.84 $\pm$ 0.36} 
& \textbf{4.39 $\pm$ 0.60} 
& \textbf{4.85 $\pm$ 0.37} \\

\bottomrule
\end{tabular}
\end{table*}

\end{document}